\documentclass[sigconf]{acmart}

\AtBeginDocument{%
  \providecommand\BibTeX{{%
    \normalfont B\kern-0.5em{\scshape i\kern-0.25em b}\kern-0.8em\TeX}}}

\usepackage{subcaption}
\usepackage{diagbox}
\usepackage{multirow}
\usepackage{multicol}
\usepackage{algorithm}
\usepackage{algorithmic}

\copyrightyear{2022}
\acmYear{2022}
\setcopyright{acmlicensed}\acmConference[KDD '22]{Proceedings of the 28th ACM SIGKDD Conference on Knowledge Discovery and Data Mining}{August 14--18, 2022}{Washington, DC, USA}
\acmBooktitle{Proceedings of the 28th ACM SIGKDD Conference on Knowledge Discovery and Data Mining (KDD '22), August 14--18, 2022, Washington, DC, USA}
\acmPrice{15.00}
\acmDOI{10.1145/3534678.3539034}
\acmISBN{978-1-4503-9385-0/22/08}

\ccsdesc[500]{Computing methodologies~Reinforcement learning}
\ccsdesc[300]{Computing methodologies~Machine learning approaches}

\keywords{Recommender System; Reinforcement Learning; Cost Modeling}

\begin{document}

\title{AutoShard: Automated Embedding Table Sharding \\ for Recommender Systems}


\author{Daochen Zha}
\email{daochen.zha@rice.edu}
\affiliation{%
  \institution{Rice University}
  \city{Houston}
  \state{TX}
  \country{USA}
}

\author{Louis Feng}
\author{Bhargav Bhushanam}
\affiliation{%
  \institution{Meta Platforms, Inc.}
  \city{Menlo Park}
  \state{CA}
  \country{USA}
}

\author{Dhruv Choudhary}
\author{Jade Nie}
\affiliation{%
  \institution{Meta Platforms, Inc.}
  \city{Menlo Park}
  \state{CA}
  \country{USA}
}

\author{Yuandong Tian}
\author{Jay Chae}
\affiliation{%
  \institution{Meta Platforms, Inc.}
  \city{Menlo Park}
  \state{CA}
  \country{USA}
}

\author{Yinbin Ma}
\author{Arun Kejariwal}
\affiliation{%
  \institution{Meta Platforms, Inc.}
  \city{Menlo Park}
  \state{CA}
  \country{USA}
}

\author{Xia Hu}
\email{xia.hu@rice.edu}
\affiliation{%
  \institution{Rice University}
  \city{Houston}
  \state{TX}
  \country{USA}
}

\renewcommand{\shortauthors}{Daochen Zha et al.}


\begin{abstract}
Embedding learning is an important technique in deep recommendation models to map categorical features to dense vectors. However, the embedding tables often demand an extremely large number of parameters, which become the storage and efficiency bottlenecks. Distributed training solutions have been adopted to partition the embedding tables into multiple devices. However, the embedding tables can easily lead to imbalances if not carefully partitioned. This is a significant design challenge of distributed systems named embedding table sharding, i.e., how we should partition the embedding tables to balance the costs across devices, which is a non-trivial task because 1) it is hard to efficiently and precisely measure the cost, and 2) the partition problem is known to be NP-hard. In this work, we introduce our novel practice in Meta, namely AutoShard, which uses a neural cost model to directly predict the multi-table costs and leverages \emph{deep reinforcement learning} to solve the partition problem. Experimental results on an open-sourced large-scale synthetic dataset and Meta's production dataset demonstrate the superiority of AutoShard over the heuristics. Moreover, the learned policy of AutoShard can transfer to sharding tasks with various numbers of tables and different ratios of the unseen tables without any fine-tuning. Furthermore, AutoShard can efficiently shard hundreds of tables in seconds. The effectiveness, transferability, and efficiency of AutoShard make it desirable for production use. Our algorithms have been deployed in Meta production environment. A prototype is available at \url{https://github.com/daochenzha/autoshard}
\end{abstract}


\maketitle

\begin{figure}
    \centering
    \includegraphics[width=0.35\textwidth]{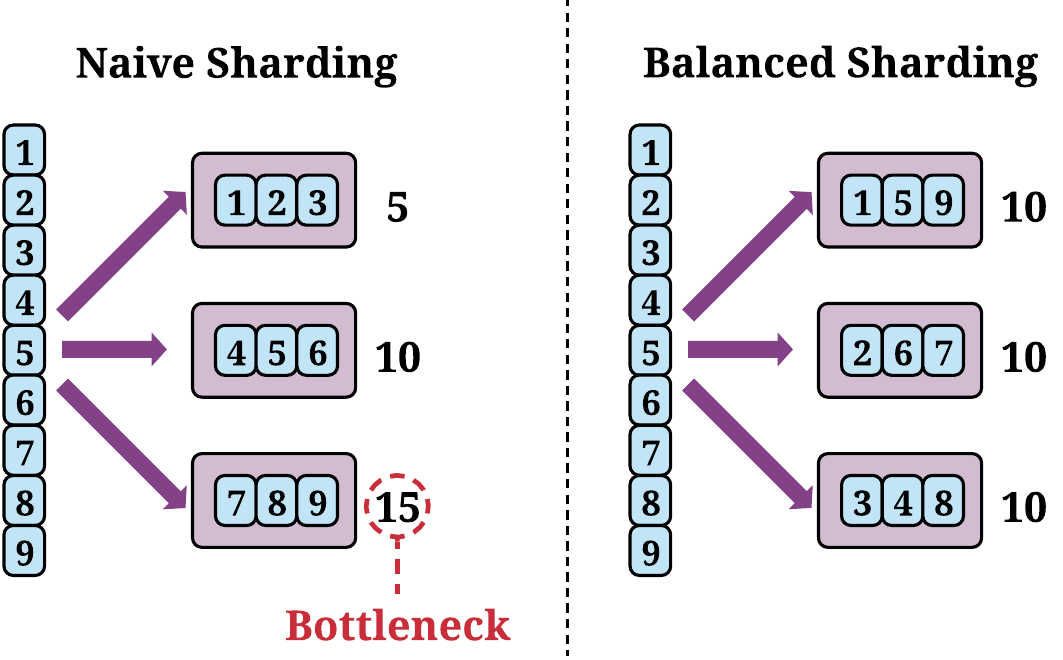}
    \vspace{-14pt}
    \caption{An illustrative sharding problem of partitioning 9 embedding tables across 3 devices with naive sharding and balanced sharding. Blue blocks are embedding tables, whose numbers indicate the costs measured by the operation execution time. Purple blocks are shards, whose costs are usually smaller than the sum of the table costs within the shard due to parallelism (e.g., 5 < 1+2+3 and 10 < 4+5+6). The slowest shard will become the bottleneck since the other shards have to wait until it finishes. Optimizing the naive sharding in this task can achieve 1.5X speedup (i.e., 15/10 = 1.5).}
    \label{fig:toysharding}
    \vspace{-10pt}
\end{figure}

\section{Introduction}

Embedding learning has become an important technique for modeling categorical features in deep recommendation models~\cite{zhang2019deep}. It maps sparse categorical features into dense vectors by performing embedding lookup in embedding tables. The learned vectors are then used for complex feature interactions and can greatly help us improve the prediction results (e.g., DeepFM~\cite{cheng2016wide}, AutoInt~\cite{song2019autoint}, and deep learning recommendation model (DLRM)~\cite{naumov2019deep}).

\begin{figure}
    \centering
    \includegraphics[width=0.47\textwidth]{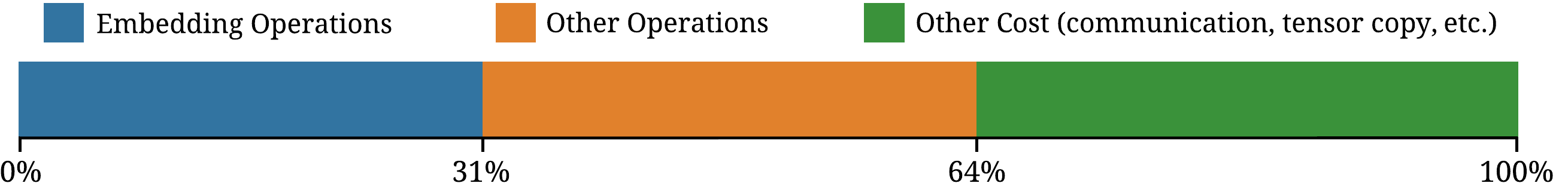}
    \vspace{-8pt}
    \caption{A breakdown of the computation time of one iteration of a Meta production model. Embedding operations account for 31\% of the iteration time and 48\% of the total operation execution time in GPU (i.e., 0.31 / 0.64 $\approx$ 48\%).}
    \vspace{-15pt}
    \label{fig:breakdown}
\end{figure}

However, industrial recommendation models often demand an extremely large number of parameters for embedding tables, which become the storage and efficiency bottlenecks~\cite{zhao2020distributed}. A typical example is YouTube Recommendation Systems~\cite{covington2016deep}, where a single categorical feature contains tens of millions of video IDs, which leads to gigantic embedding tables. The ultra-large embedding tables also result in training efficiency problems. For instance, more than 48\% of the operation kernel time is spent on embedding tables in a Meta production model (see Figure~\ref{fig:breakdown} for the breakdown). Similar observations are also reported in~\cite{acun2021understanding}, showing that embedding table sizes have a significant impact on the training throughput. The memory and efficiency requirements motivate the distributed training solutions, where model-parallelism is exploited to partition and feed the embedding tables into multiple devices~\cite{acun2021understanding,naumov2019deep,zhao2020distributed,amazon-dsstne,gupta2020architectural,naumov2020deep}. The embedding lookup for a certain index will then be performed by querying the device that actually holds the corresponding table.


While the model-parallelism enables training models with very large embedding sizes, it poses a significant design challenge named \emph{embedding table sharding}, i.e., how we should partition the embedding tables across devices. Figure~\ref{fig:toysharding} presents an illustrative example of why optimizing the sharding can significantly accelerate the training. If not carefully partitioned (left-hand side of Figure~\ref{fig:toysharding}), the tables could lead to imbalances among GPUs, where all the GPUs are forced to wait for the slowest GPU\footnote{In this work, we focus on embedding table sharding among GPU devices when the tables fit on the GPU memory. We note that system memory~\cite{jiang2019xdl}, non-volatile memory~\cite{jeong2018nonvolatile}, and SSD~\cite{zhao2020distributed} can also be used to store tables at the cost of decreased throughput~\cite{acun2021understanding}. We defer hybrid sharding strategies for these scenarios to future work.}. In contrast, a balanced sharding (right-hand side of Figure~\ref{fig:toysharding}) can significantly accelerate the embedding operation by reducing the waiting time. Motivated by this, we investigate the following research question: \emph{given an arbitrary set of embedding tables, how can we shard the embedding tables to balance the costs across 
devices?}

It is non-trivial to achieve this goal because of two major challenges. \textbf{First}, we need to efficiently estimate the cost (i.e., the kernel time of the embedding operators), which serves as the optimization objective. Unfortunately, the cost is hard to estimate. Unlike many other partition problem, the total cost of multiple tables in a shard is not the sum of the single table costs within the shard due to parallelism and operator fusion. As a result, it is hard to estimate the cost without actually running the operators; however, running the operators is computationally expensive. \textbf{Second}, we need an efficient algorithm to solve the partition problem, which is known to be NP-hard\footnote{\url{https://en.wikipedia.org/wiki/Partition_problem}}. The ever-increasing number of embedding tables makes it infeasible to adopt a brute-force approach, i.e., iterating through all the possible sharding plans and outputting the best one. For practical use, the sharding algorithm is expected to propose an effective sharding plan for hundreds of tables in a reasonable time.

To address the above challenges, we present our novel practice in Meta, namely AutoShard, based on cost modeling and deep reinforcement learning (RL). To efficiently estimate the cost, we develop a neural cost model, which is trained with regression to the multi-table cost data collected from running micro-benchmarks on GPUs. To optimize the partitioning, we formulate the table sharding as a Markov decision process (MDP), where in each step, we allocate one table to a shard. The process ends when all the tables are allocated and we obtain a reward indicating the sharding quality at the final step. Then we leverage deep RL to learn an LSTM policy to optimize the sharding strategy. The cost model and the sharding policy are jointly trained towards convergence. In summary, we make the following contributions:





\begin{itemize}
    \item Provide an in-depth analysis of the main influential factors of the cost of embedding operators via a case study on an modern embedding bag implementation from FBGEMM~\cite{fbgemm}.
    \item Propose AutoShard for embedding table sharding. It uses neural cost model to predict the kernel time of the operator and leverages deep RL to solve the partition problem. It can propose an effective sharding plan for hundreds of tables in seconds with a single CPU core.
    \item Conduct extensive experiments on an open-sourced large-scale synthetic dataset (for reproducibility) and Meta's production dataset. AutoShard significantly outperforms the heuristic sharding strategies. In particular, AutoShard can well transfer to various scenarios. The trained policy of AutoShard can be directly applied to solve a wide range of sharding tasks with various numbers of tables and different ratios of the unseen tables without any fine-tuning, achieving the same level of balance and speedup. 
    
\end{itemize}

\begin{figure}
    \centering
    \includegraphics[width=0.42\textwidth]{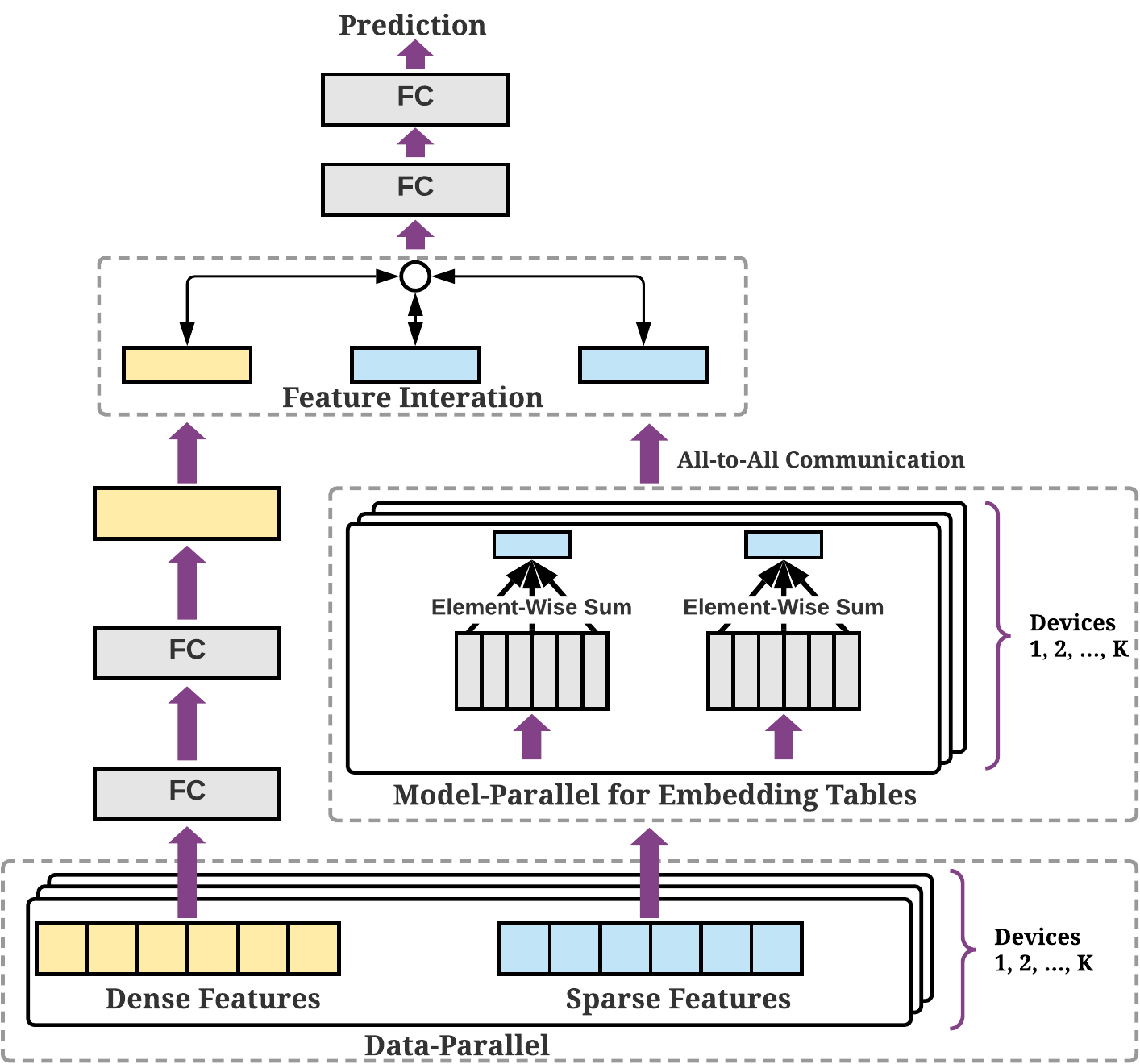}
    \vspace{-5pt}
    \caption{A typical recommendation model with dense and sparse features~\cite{naumov2019deep}. The system exploits a combination of model parallelism (i.e., the embedding tables are partitioned into different devices) and data parallelism (i.e., replicating MLPs on each device and partitioning training data into different devices). The embedding vectors obtained from embedding lookup are appropriately sliced and transferred to the target devices through an all-to-all communication. }
    \vspace{-15pt}
    \label{fig:dlrm}
\end{figure}

\begin{figure*}[t]
  \centering

  \begin{subfigure}[b]{0.33\textwidth}
    \centering
    \includegraphics[width=0.8\textwidth]{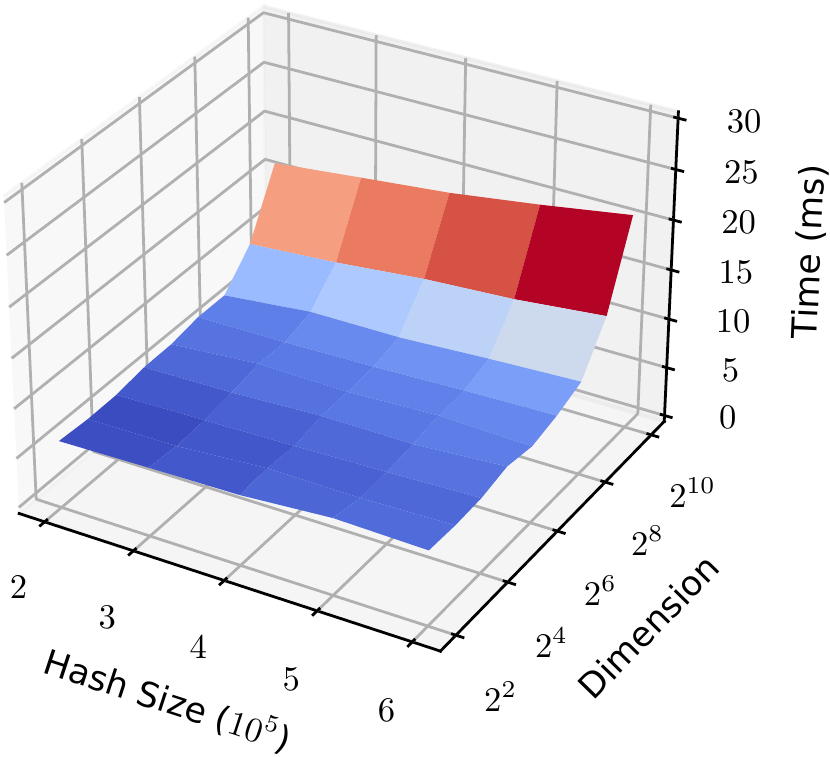}
    \vspace{-8pt}
  \end{subfigure}%
  \begin{subfigure}[b]{0.33\textwidth}
    \centering
    \includegraphics[width=0.8\textwidth]{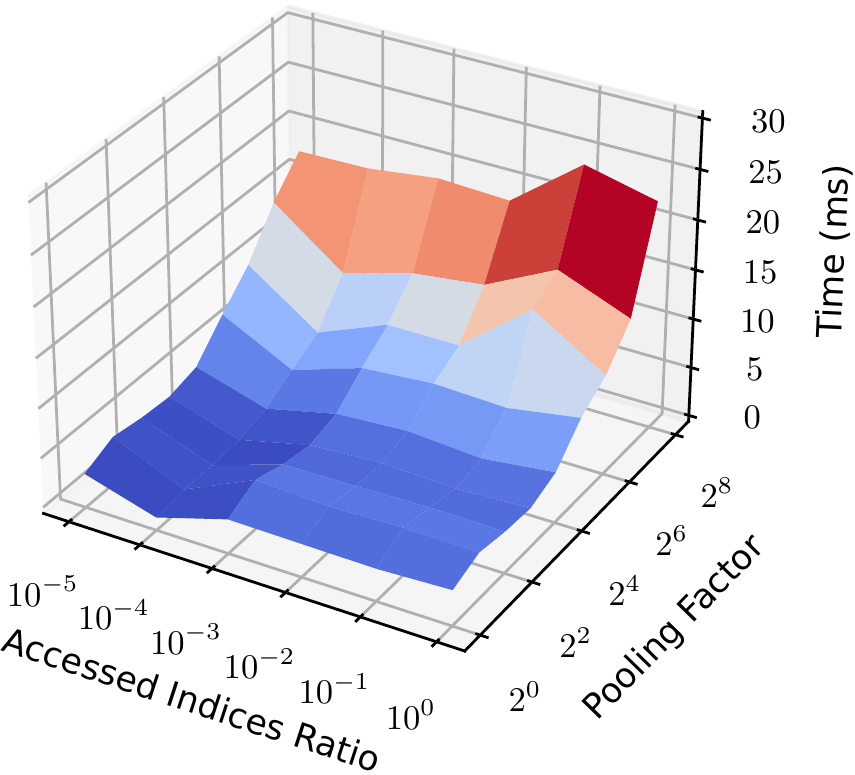}
    \vspace{-8pt}
  \end{subfigure}%
  \begin{subfigure}[b]{0.34\textwidth}
    \centering
    \includegraphics[width=0.85\textwidth]{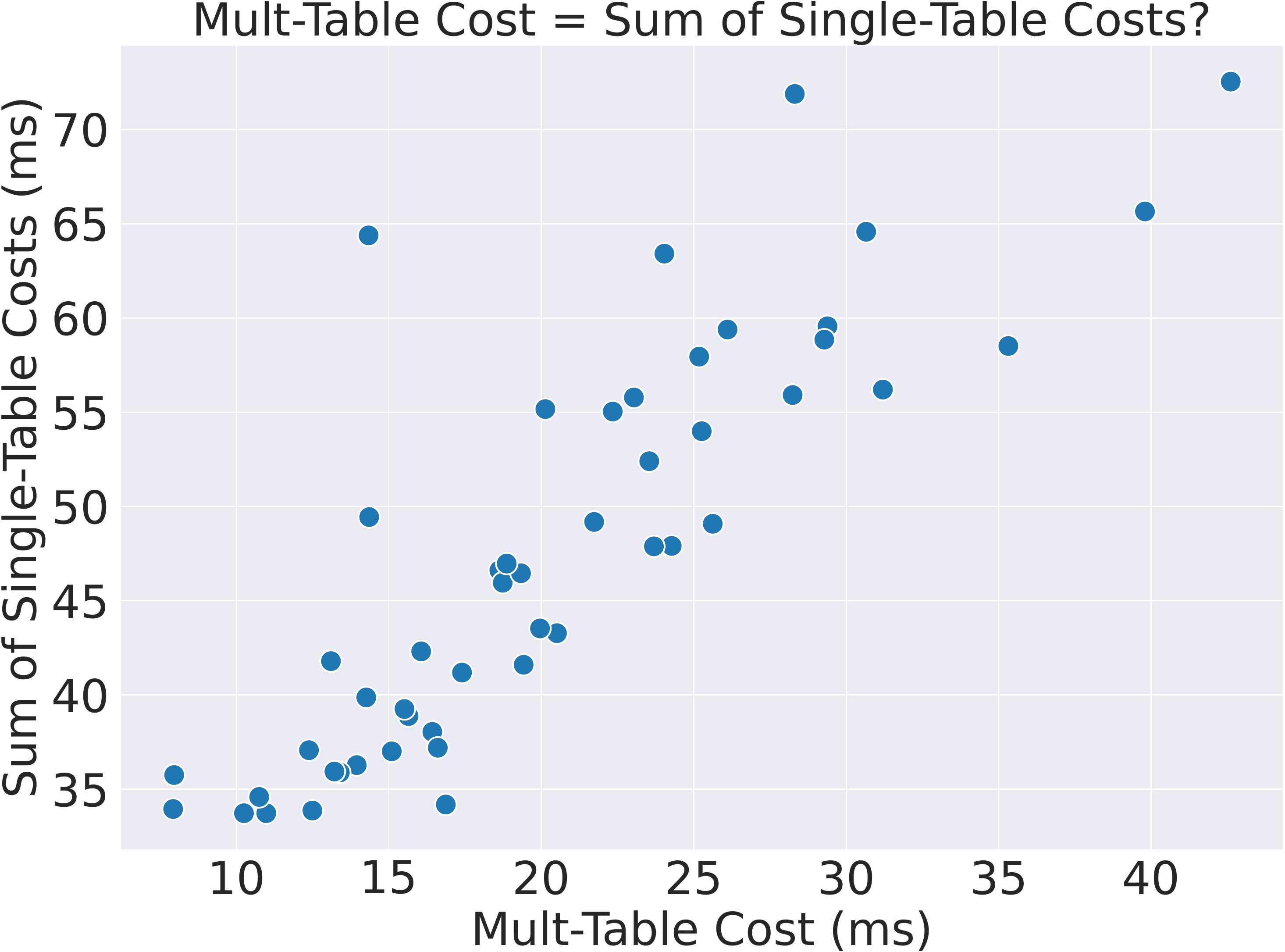}
    \vspace{-8pt}
  \end{subfigure}%
  \caption{Impact of hash size and dimension on the single table cost (left); impact of indices distribution and pooling factor on the single table cost (middle); multi-table cost versus the sum of single table costs (right).}
  \vspace{-8pt}
  \label{fig:analysis}
\end{figure*}

\section{Preliminaries}

We start with a background of distributed recommender systems, and then formulate the embedding table sharding problem.

\subsection{Training Deep Learning Recommendation Models at Scale}
\label{sec:21}
Industrial recommendation models usually require massive memory and high training throughput. Therefore, distributed training is employed in large-scale industrial systems~\cite{acun2021understanding,amazon-dsstne,covington2016deep,zhou2019deep,liu2017related,gomez2015netflix}. We take DLRM\footnote{\url{https://github.com/facebookresearch/dlrm}}~\cite{naumov2020deep} as an example to introduce the system design.

Figure~\ref{fig:dlrm} depicts an overview of DLRM. The system exhibits a combination of data parallelism and model parallelism. To exploit data parallelism, each trainer will hold a copy of MLP layers, which will be trained on its own mini-batch of data. The model parameters are updated in the fully synchronous mode. For embedding tables, model parallelism is adopted to shard the tables into multiple devices. In the forward pass of embedding lookup, each device will perform a lookup for all the indices concerning the tables in the device (including the indices from the other devices' mini-batches). The obtained vectors will be transferred to the corresponding devices via an all-to-all communication. Then each device will receive all the embedding vectors for its own mini-batch, which will be interacted with the dense features, followed by an MLP layer for predictions. In the backward pass, the gradients will be similarly transferred with another all-to-all communication so that each device will receive all the gradients for the tables within the device.




\subsection{Problem Formulation}

In the production environment, the possible embedding tables will often stay unchanged for a period of time since the raw features for a recommendation task are relatively steady. Nevertheless, different subsets of the tables could be used to build recommendation models. For example, a machine learning engineer may conduct feature selection by experimenting with different table combinations.


Following the above intuition, we formalize the embedding table sharding problem. Let $\mathcal{T}_p = \{T_1, T_2, ..., T_N\}$ be a pool of embedding tables, where $N$ is the total number of tables in the pool. A sharding task $S$ can be represented as a triple $S = (\mathcal{T}, \mathcal{D}, \mathcal{M})$, where $\mathcal{T} \subseteq \mathcal{T}_p$ is subset of the tables, $\mathcal{D} = \{1, 2, ..., K\}$ is a set of shard IDs with $K$ shards in total, and $\mathcal{M}=\{M_1, M_2, ..., M_K\}$ is the memory constraints for all the shards. A sharding plan $\pi$ can be represented as a mapping from each table to a shard. Then each device will get its own shard to process, which leads to a set of actual memory usages $\hat{\mathcal{M}}=\{\hat{M}_1, \hat{M}_2, ..., \hat{M}_K\}$ (which is obtained by summing the tables sizes in each shard) and a set of costs $\mathcal{C} = \{C_1, C_2, ... C_K\}$ in terms of operation execution time. Embedding table sharding aims to optimize the sharding plan $\pi$ such that the maximum cost across shards is minimized subject to the memory constraints:

\vspace{-10pt}
\begin{equation}
\begin{aligned}
\min_{\pi} \quad \max(\mathcal{C}) := \max_k C_k \quad \textrm{s.t.} \quad \hat{M}_k \leq M_k, \forall k \in \mathcal{D}.   \\
\end{aligned}
\end{equation}
\vspace{-15pt}




\section{Analysis of Embedding Table Cost}

This section analyzes the table costs on a modern embedding bag implementation from FBGEMM\footnote{\url{https://github.com/pytorch/FBGEMM/}}~\cite{fbgemm} with a 2080Ti GPU.


\subsection{Analysis of Single-Table Cost} The cost of a table is mainly determined by the characteristics of the table itself and indices lookup. Table characteristics include \emph{hash size}, which is defined as the number of entries of the table, and \emph{dimension}, which means the dimension of the embedding vectors. Characteristics of indices lookup include \emph{pooling factor}, which is the average number of lookup indices per query, and \emph{indices distribution}, which determines the indices accessing frequencies. We study the table characteristics and indices lookup on synthetic tables.

We first visualize the impact of table characteristics. We fix the pooling factor to be 32 and the indices to be uniformly distributed. Then we vary the hash size and table dimension to plot the kernel time in the left-hand side of Figure~\ref{fig:analysis}. As expected, we can see that a higher dimension will significantly increase the kernel time. This is because the dimension is positively correlated to the amount of data to be fetched. An interesting observation is that the hash size only has a moderate impact on the cost, which could be partly explained by the $O(1)$ time complexity for hash table lookup.

Similarly, we study the impact of indices lookup by fixing hash size to be $10^6$ and dimension to be 32. For the indices distribution, some indices could be accessed far more frequently than others~\cite{acun2021understanding}. We simulate this behavior by restricting the indices access to a subset of the indices with a pre-defined accessed indices ratio. For example, a ratio of 1.0 is equivalent to the uniform distribution, while a ratio of $10^{-2}$ suggests only one percent of the indices will be accessed, which means the indices are sparsely distributed and those one percent of the indices are warm indices. The middle of Figure~\ref{fig:analysis} shows the impact of the pooling factor and accessed indices ratio. A larger pooling factor will significantly increase the time, which is expected since it indicates more indices per lookup. In contrast, a sparse indices distribution tends to decrease the time. We speculate that this is caused by the caching mechanism.

\begin{figure}
    \centering
    \includegraphics[width=0.37\textwidth]{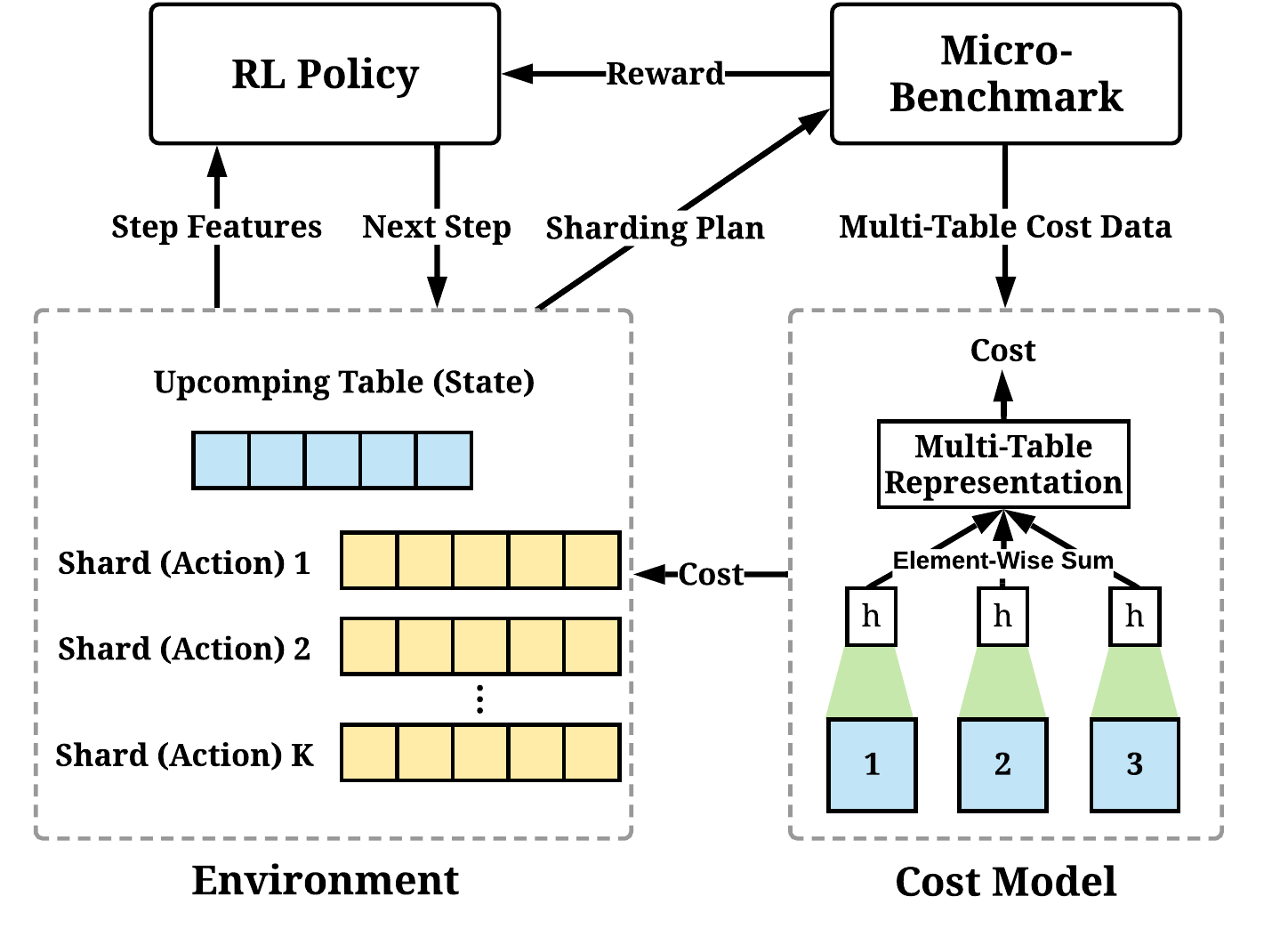}
    \vspace{-12pt}
    \caption{An overview of AutoShard. The RL policy interacts with the environment and allocates embedding tables one by one. Based on the generated shards, the micro-benchmark measures the actual latencies of embedding table operators and produces a reward to update the RL policy. The cost model approximates multi-table costs based on the data collected from the micro-benchmark. The estimated costs will be used to enhance the action representations in RL training.}
    \vspace{-18pt}
    \label{fig:overview}
\end{figure}

The above analysis motivates a series of greedy sharding algorithms, which will be elaborated in Section~\ref{sec:51}. However, the designed heuristics are sub-optimal since the actual running time has non-linear and complex relationships with these four factors.



\vspace{-3pt}
\subsection{Analysis of Multi-Table Cost}
\label{sec:232}

Since we usually have multiple tables in a shard, we need to estimate the multi-table cost. A naive way is to sum the costs of the single tables within the shard (single-sum for short). However, this is inaccurate due to the parallelism of GPU. The right-hand side of Figure~\ref{fig:analysis} plots the multi-table cost versus the single-sum of 50 randomly sampled table combinations from MetaSyn (see Section~\ref{sec:51} for dataset details), where each sample contains 10 tables. \textbf{First}, we observe that the multi-table cost is significantly smaller than the single-sum. This is because the tables can be batched and accelerated with parallelism. \textbf{Second}, while the single-sum is positively correlated with the multi-table cost in general, it is still a poor estimator in many cases. The discrepancy between the multi-table cost and single-sum will amplify the difficulty of cost estimation.



\vspace{-3pt}
\section{Methodology}
An overview of AutoShard is shown in Figure~\ref{fig:overview}. It consists of four modules: \emph{1) a micro-benchmark} that measures actual costs of embedding operators~(Section~\ref{sec:31}), \emph{2) a cost model} which approximates multi-table costs based on the data collected from micro-benchmark~(Section~\ref{sec:32}), \emph{3) an environment} that formulates the sharding process as a Markov Decision Process~(MDP) by allocating one table in each step~(Section~\ref{sec:33}), and \emph{4) an RL policy} that optimizes the sharding strategies in a trial-and-error fashion~(Section~\ref{sec:34}). Finally, we summarize the training procedure in Section~\ref{sec:35}.


\vspace{-3pt}
\subsection{Micro-Benchmarking Embedding Operators}
\label{sec:31}
This subsection introduces how we efficiently and precisely measure the latency of embedding tables. Since we only need the latency of embedding operators, we design and implement a micro-benchmark, which only benchmarks embedding operators alone. Due to space limitation, we introduce the main steps of micro-benchmarking here and provide more details in Appendix~\ref{sec:appendix21}: \emph{1) initialization:} we initialize the operators with the specified arguments of embedding tables and load the indices data. \emph{2) warmup}: we run the embedding operator several times to warm up the device to allow the CUDA to complete the necessary preparation for the operator. \emph{3) benchmarking}: run embedding operator several times and return the mean latency\footnote{The micro-benchmark is currently maintained as a separate open-source effort for all the PyTorch operators at \url{https://github.com/facebookresearch/param}}.

\begin{figure*}
    \centering
    \includegraphics[width=0.85\textwidth]{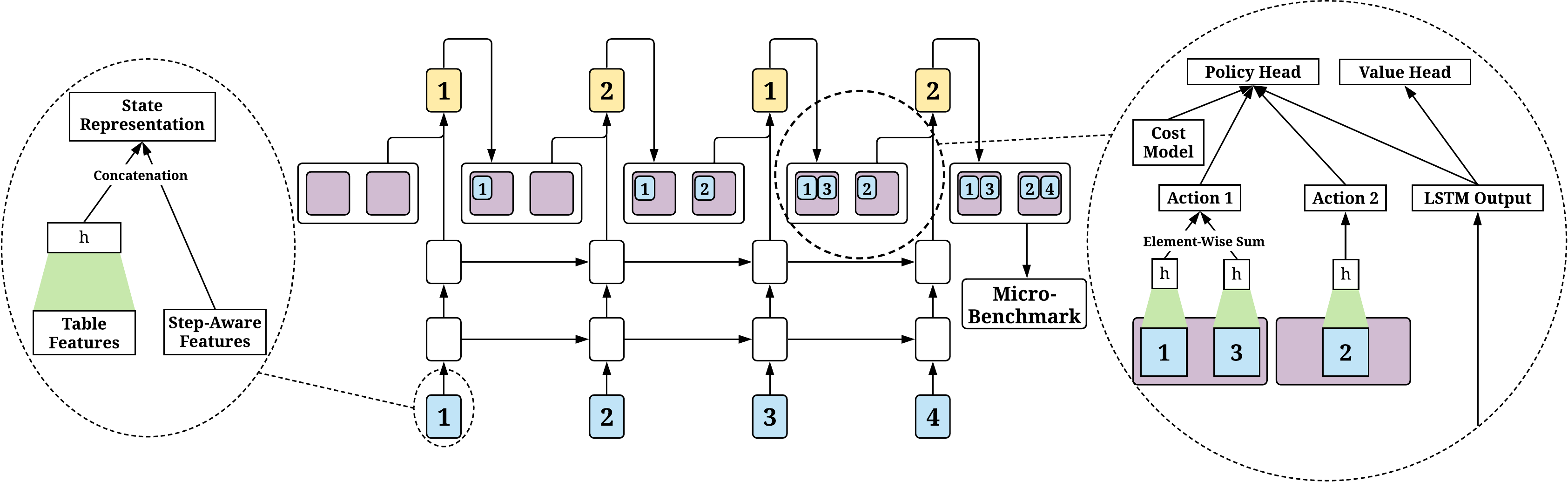}
    \vspace{-12pt}
    \caption{An illustration of the sharding process of AutoShard. A two-layer LSTM encodes historical state representations and in each step outputs a shard ID (yellow) that allocates the table at hand (blue) to a shard (purple). State representation is obtained by concatenating the table representation and step-aware features (dotted circle in the left). Action representation is the sum of the single tables' representations (dotted circle in the right). A policy head produces a shard ID based on the state/action representations and the cost model. A value head will approximate state values to reduce the variance of RL training.}
    \vspace{-9pt}
    \label{fig:arch}
\end{figure*}

\subsection{Modeling Multi-Table Costs}
\label{sec:32}
While micro-benchmark can accurately and efficiently measure the latency, it still needs to run the operators, which remains computationally expensive for production use. To further accelerate the cost estimation, we develop a neural cost model to predict the operator cost based on the data collected from the micro-benchmark. The neural cost model can be easily deployed since it only requires a forward pass of a shallow neural network to obtain the cost.

Cost estimation can be formulated as a regression task, which takes as input the features of multiple tables and outputs the latency. The lower right corner of Figure~\ref{fig:overview} illustrates the neural architecture of the cost model. For each table, we use a shared MLP (green) to generate the single table representations based on the table features. Then we sum up the single table representations to obtain a multi-table representation (we have tried other reductions, such as max and mean, and found that sum works better), followed by another MLP to predict the cost. This design can flexibly accommodate different numbers of tables and obtain a final representation with a fixed dimension. We empirically use the following features: table dimension, hash size, pooling factor, table size, and indices distributions (17 features). We provide more details of these features in Appendix~\ref{sec:appendix22}. Formally, let $(\textbf{X}, \textbf{y})$ be the collected data. $\textbf{X}$ is the table features where each row represents the features of multiple tables and has variable lengths; $\textbf{y}$ denotes a vector of ground-truth costs obtained by running micro-benchmark on GPUs. We train the cost model $f$ with mean squared error (MSE) loss $L_\text{cost} = (\textbf{y} - f(\textbf{X}))^2$. The performance of the cost model is reported in Table~\ref{tbl:costmodel} of Appendix.

\subsection{Formulating Sharding as MDP}
\label{sec:33}
This subsection describes why and how we formulate the sharding procedure as a sequential decision process. A naive strategy to solve the sharding problem is to treat it as a black-box optimization problem, where we sample and evaluate a sharding plan in each iteration. However, this will lead to an extremely large search space since each table can be possibly assigned to any shard.

To tackle this problem, we decompose the generation of a sharding plan into multiple steps, where we only assign one table to a shard in each step. Then after scanning all the tables, we can eventually obtain a sharding plan. This formulation has two desirable properties. \textbf{First,} it can significantly decrease the decision space. Specifically, the decision space of the sharding policy is only the number of shards $K$ in each step. Although the policy needs to perform more steps, the decisions made across steps are very similar (i.e., they all aim to assign a table to achieve load balance) so that the policy in one step may learn to reuse the knowledge learned from other steps and improve learning efficiency. \textbf{Second,} it can implicitly encourage transferable strategies. By associating one table with one step, a model trained on very few tables can easily transfer to more tables by simply adding more steps without re-training. This cannot be achieved by black-box optimization.

We formulate the above process as an MDP with the state, action, and reward defined as follows. \textbf{State:} The features of the upcoming table and a step-aware feature which is the ratio of the remaining tables to be assigned. \textbf{Action:} The shard IDs with $K$ actions in total. The multi-table cost features and the predicted costs from the cost model can serve as the action features. \textbf{Reward}: The agent will receive zero rewards for all the intermediate steps and a final reward indicating the quality of the sharding plan. Specifically, if the sharding plan meets the memory constraint, we run the micro-benchmark to obtain the shard latencies of all the shards. The reward is calculated by the ratio between the maximum latency and minimum latency, i.e., $min(\mathcal{C})/max(\mathcal{C})$, to encourage the agent to balance the costs across shards. The reward is in the range of $[0,1]$, where a higher reward suggests a better balance. Alternatively, if the sharding plan cannot meet the memory constraints, we penalize this behavior with a negative reward, which is determined by the shard that is the most seriously affected by the memory explosion, i.e.,  $\max_k ((\hat{M}_k - M_k) / M_k)$. A higher reward indicates that the sharding plan is closer to meeting the memory constraints.

\vspace{-9pt}

\begin{figure}
    \centering
    \includegraphics[width=0.27\textwidth]{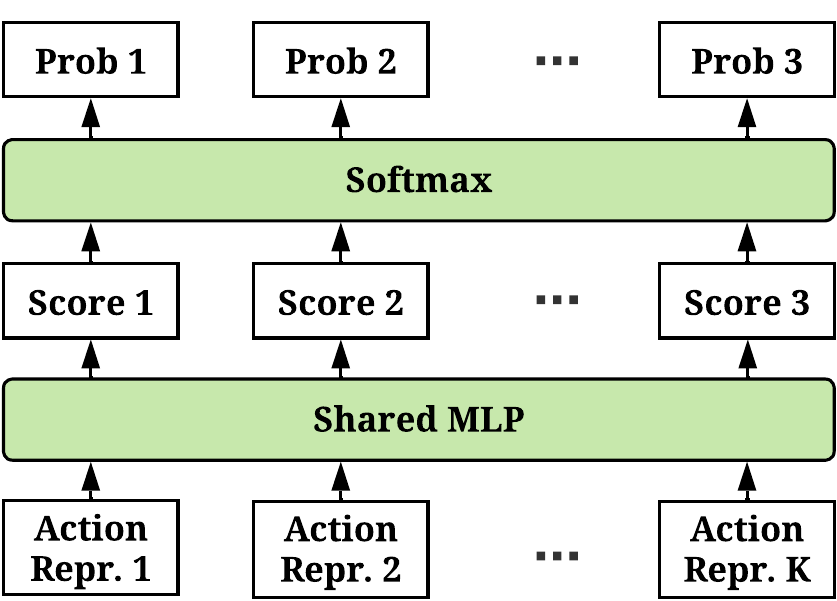}
    \vspace{-8pt}
    \caption{An illustration of the policy head.}
    \vspace{-20pt}
    \label{fig:actionconfidence}
\end{figure}

\subsection{Optimizing Sharding Strategy with RL}
\label{sec:34}
This subsection presents how we solve the above MDP with RL. We will first elaborate on the neural architecture design and then describe how we train the model weights.

Figure~\ref{fig:arch} illustrates the neural architecture and how it makes predictions to shard tables. The model takes as input the state features and actions features, and outputs an action probability vector where each action corresponds to a shard ID (policy head) and a scalar value indicating the value of the state (value head). The model is instantiated with a two-layer LSTM to process the state and actions with the following procedure. \textbf{1)} The state representation is obtained by combining the table features and step-aware features (dotted circle in the left). \textbf{2)} The state representations will be fed into the LSTM sequentially so that historical state information will be encoded into LSTM output as well. \textbf{3)} The multi-table representations obtained in the cost model will be concatenated with the predicted costs from the cost model to construct the action representations (dotted circle in the right). \textbf{4)} Each action representation will be concatenated with the state representation, followed by an MLP (which is shared for all actions) to produce a confidence score for the action, shown in Figure~\ref{fig:actionconfidence}. \textbf{5)} The scores for all the actions will be processed by a Softmax layer to obtain the probability vector, where the probabilities sum to one. Similarly, the LSTM output will be fed into another MLP to generate a state value in the value head. For an upcoming table, the model will sample a shard ID (action) based on the probabilities from the policy head. The tables will be allocated to the shards one by one following this procedure.

We train the model with IMPALA~\cite{espeholt2018impala}, a distributed actor-critic algorithm enhanced by V-trace targets to achieve off-policy learning. The V-trace correction can tackle delayed model weights update, which is helpful in our problem in that evaluating a sharding plan is slow and may result in substantial delays. Here, we only briefly introduce IMPALA since RL itself is not our focus; one can adopt other RL algorithms as well under our framework. We first introduce the V-trace targets and then describe the loss for updating the policy and value heads. Let $s_t$, $a_t$, and $r_t$ be the state, action, and reward at step $t$, respectively. We consider an n-step trajectory $(s_t, a_t, r_t)_{t=t'}^{t=t'+n}$. The V-trace target for $s_{t'}$ is defined as
\begin{equation}
    V_{\text{target}}(s_{t'}) = V(s_{t'}) + \sum_{t=t'}^{t'+n-1} \gamma^{t-t'}(\pi_{i=t'}^{t-1} c_i) \delta_t V,
    \label{eq:1}
\end{equation}
where $V(s_{t'})$ is the output of the value head for $s_{t'}$, $\delta_t V = \rho_t (r_t + \gamma V(s_{t+1}) - V(x_t))$ is the temporal difference, and $c_i$ and $\rho_t$ are truncated importance sampling weights that tackle the delayed update of the model. Then the loss at step $t$ is defined as
\begin{equation}
    L_t = \rho_t \log \pi(a_t|s_t) (r_t + \gamma V_{\text{target}}(s_{t+1} - V(s_t)) + \frac{1}{2} (v_t - V(s_t))^2,
    \label{eq:2}
\end{equation}
where $\pi(a_t|s_t)$ and $V(s_t)$ correspond to policy and value heads, respectively. The training can be batched to update the losses for multiple steps at a time in each iteration.

\subsection{Training of AutoShard}
\label{sec:35}
This subsection summarizes the overall training procedure. To improve the sample efficiency, the cost model and the LSTM policy-value network are jointly trained with shared table representations and data. Specifically, the MLP for processing the tables features (i.e., the green parts in Figure~\ref{fig:overview} and Figure~\ref{fig:arch}) is shared. Similarly, the data collected by the RL agent will be reused to generate cost data to train the cost model. The whole training process is summarized in Algorithm~\ref{alg:1}. In each training iteration, we collect a batch of trajectories by interacting with the micro-benchmark to update the policy-value network (line 4). The collected data will be stored in a buffer and reused to train the cost model (line 8). Since the main bottleneck is data collection, we parallelize line 4 with multiple processes operating on different GPUs. Once the cost model and the policy-value network are trained, they can be directly applied to any new sharding tasks by sequentially predicting the shard IDs.

\section{Experiments}
The experiments are conducted on both synthetic datasets and production datasets at Meta. We aim to answer the following questions: \textbf{Q1:} How does AutoShard compare with the heuristic sharding strategies~(Section~\ref{sec:42})? \textbf{Q2:} How large is the search space of AutoShard and can simple random search be competitive with it~(Section~\ref{sec:43})? \textbf{Q3:} How does each component of AutoShard contribute to the performance~(Section~\ref{sec:44})? \textbf{Q4:} Can AutoShard transfer to unseen tables, and sharding tasks with more tables~(Section~\ref{sec:45})? \textbf{Q5:} How efficient is the training/inference of AutoShard~(Section~\ref{sec:46})?


\begin{algorithm}[t]
\caption{Training of AutoShard}
\label{alg:1}
\setlength{\intextsep}{0pt} 
\begin{algorithmic}[1]
\STATE \textbf{Input:} Training tasks $\mathcal{S}_{\text{train}} = \{S_i\}_{i=1}^{n}$, batch size $B_1$ for policy-value network and $B_2$ for cost model, number of update iterations $I$ for cost model, number of data collection steps $T$
\STATE Initialize the cost model and policy-value network
\FOR{iteration = 1, 2, ... until convergence}
    \STATE Collect a set of trajectories with $T$ steps $\{s_t, a_t, r_t\}_{t=1}^{T}$ from a randomly sampled task from $\mathcal{S}_\text{train}$ and store the generated cost data into a buffer
    \IF{more than $B_1$ sets of new trajectories are collected}
        \STATE Update policy-value network with Eq.~\ref{eq:2}
        \FOR{iteration = 1, 2, ..., $I$}
            \STATE Sample a batch of cost data with size $B_2$ from the buffer and update the cost model with MSE loss
        \ENDFOR
    \ENDIF
\ENDFOR
\end{algorithmic}
\end{algorithm}

\subsection{Experimental Settings}
\label{sec:51}

\textbf{Datasets.} The public recommendation datasets often cannot match the scale of the industrial models to enable a valid evaluation of embedding table sharding. For example, Criteo\footnote{https://www.kaggle.com/c/avazu-ctr-prediction/data}, one of the most popular datasets, only has 26 sparse features with a cardinality of at most one million. Thus, our experiments are mainly conducted on an open-sourced large-scale synthetic dataset\footnote{\url{https://github.com/facebookresearch/dlrm_datasets}} (\textbf{MetaSyn}), which shares similar indices accessing patterns to Meta production embedding tables. As tabulated in Table~\ref{tab:stats}, MetaSyn consists of hundreds of embedding tables with very large and diverse hash sizes and pooling factors. Since MetaSyn does not specify table dimensions, we randomly select a dimension for each table from $\{16, 32\}$ (we purposely make the dimensions small so that our results can be reproduced on GPUs with 10 GB memory). For verification, we also conduct experiments on the Meta production embedding tables (\textbf{MetaProd}), which have a similar scale as $\text{MetaSyn}$ except for larger table dimensions. We keep the details of MetaProd confidential. MetaSyn and MetaProd will serve as the table pools, where each sharding task is constructed by a randomly sampled subset of the tables. Intuitively, with more tables, we also need more devices so that the tables can fit on the GPU memory. We empirically set the number of devices to be 1/10 of the total number of tables for all the tasks since this setting can ensure sufficient memory for all the sharding algorithms on both MetaSyn and MetaProd; that is, we use 8 devices for 80 tables, and 16 devices for 160 tables, etc. We provide more details in Appendix~\ref{sec:appendix1}.

\textbf{Heuristic Algorithms.} We compare AutoShard against several deployed heuristic sharding algorithms. They mainly consist of two steps: \emph{(1) cost function:} each table will be assigned an estimated cost, and \emph{(2) greedy algorithm:} the tables are first sorted in descending order based on the costs. Starting from the table with the highest cost, greedy algorithm will assign tables one-by-one to the device with the lowest sum of the costs so far, so that each device will have roughly an equal sum of the costs in the end. We consider the heuristics with the following cost functions, which have been proven to show strong performance in prior work~\cite{sethi2022recshard}: the size (the product of dimension and hash size) of the table (\textbf{size-greedy}), the dimension of the table (\textbf{dim-greedy}), the product of the dimension and mean pooling factor of the table (\textbf{lookup-greedy}). We further include a random sharding baseline (\textbf{rand}).

\textbf{Metrics.} We evaluate the performance with the following metrics. \textbf{Degree of balance:} the ratio between the minimum latency and the maximum latency across shards. 100\% suggests perfect balancing where each shard has equal latency, and 0\% indicates the worst-case of load balance. \textbf{Speedup:} the speedup over random sharding which is the most naive strategy. Specifically, the speedup is calculated by $max(\mathcal{C}^{random})/max(\mathcal{C})$, where $\mathcal{C}^{random}$ and $\mathcal{C}$ are the sets of actual embedding table costs of random sharding strategy and the sharding algorithm at hand, respectively. 

\textbf{Implementation Details.} All the hyperparameters are tuned based on MetaSyn and the same set of hyperparameters are used on MetaProd. Specifically, we set $B_1=8$, $B_2=512$, $I=20$, and $T=100$. We use the IMPALA implementation in~\cite{kuttler2019torchbeast} with the default hyperparameters for RL training. We use 2080 Ti and V100 GPUs for MetaSyn and MetaProd, respectively. We run all the experiments five times with random seeds 0, 1, 2, 3, and 4 and report the means and standard deviations. We provide more details in Appendix~\ref{sec:appendix3}.




\begin{table}[]
    \centering
    \small
    \caption{Statistics of Meta synthetic embedding tables.}
    \vspace{-8pt}
    \label{tab:stats}
    \begin{tabular}{l|l}
    \toprule
    \textbf{Attribute} & \textbf{Value} \\
    \midrule
     Number of Tables & 856\\
     Batch Size & 65,536\\
     Max/Mean/Min Hash Sizes & 12,543,670 / 4,107,458 / 1\\
     Max/Mean/Min Pooling Factors & 193 / 15 / 0\\
    \bottomrule
    \end{tabular}
    \vspace{-8pt}
\end{table}

\begin{figure}[t]
  \centering
  \begin{subfigure}[b]{0.46\textwidth}
    \centering
    \includegraphics[width=0.99\textwidth]{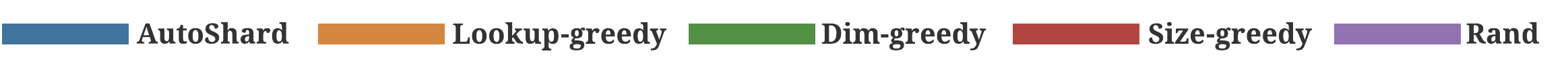}
  \end{subfigure}%
 
  \begin{subfigure}[b]{0.23\textwidth}
    \centering
    \includegraphics[width=0.99\textwidth]{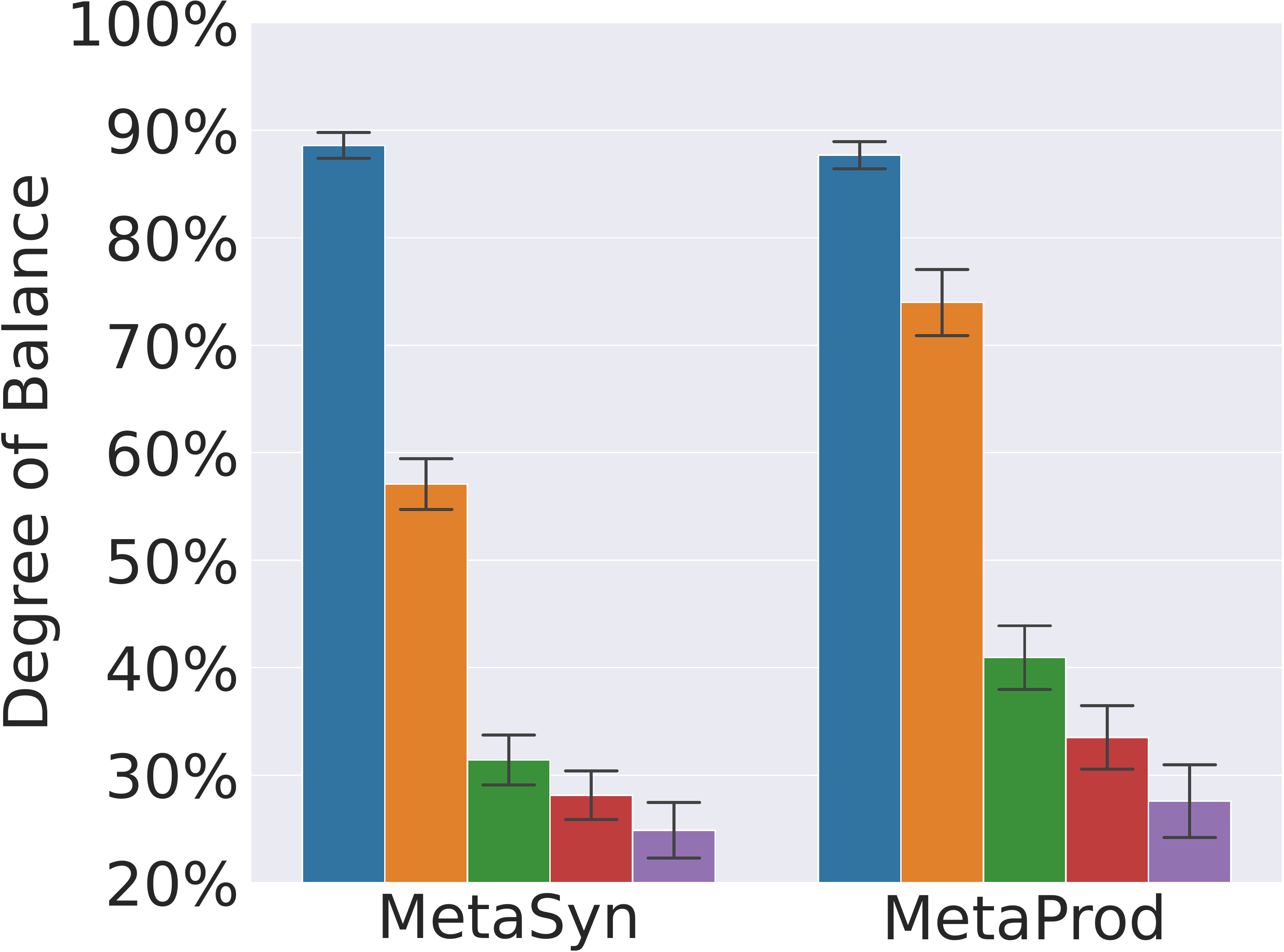}
    \caption{Degree of balance}
    \vspace{-10pt}
  \end{subfigure}%
  \begin{subfigure}[b]{0.23\textwidth}
    \centering
    \includegraphics[width=0.99\textwidth]{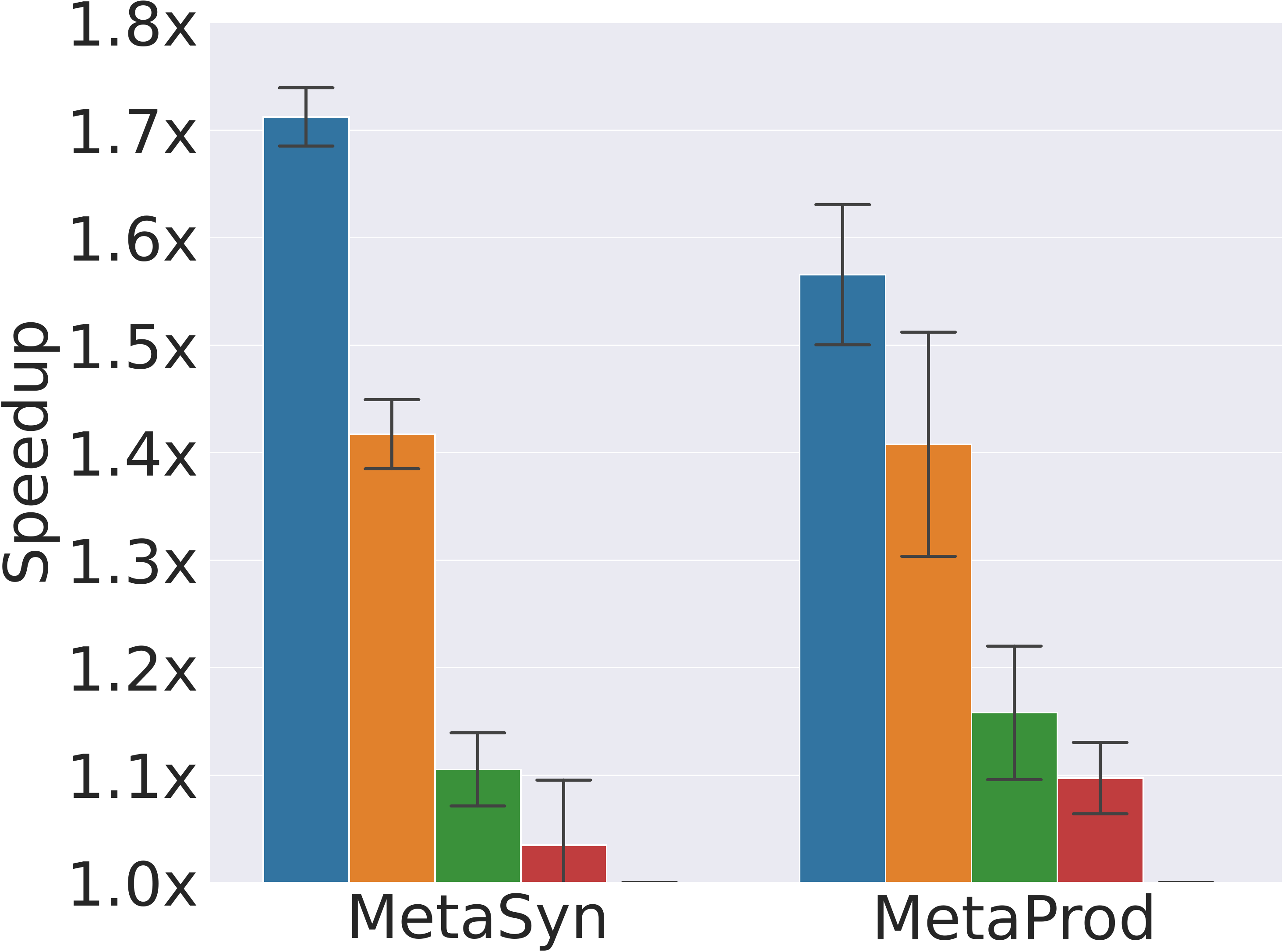}
    \caption{Speedup}
    \vspace{-10pt}
  \end{subfigure}%
  \caption{Performance of AutoShard against baselines. We report the mean and standard deviation across five runs.}
  \vspace{-18pt}
  \label{fig:compare}
\end{figure}

\subsection{Comparison with the Heuristics}
\label{sec:42}

To study \textbf{Q1}, we conduct experiments on the tasks of sharding 80 tables to 8 devices. Specifically, we randomly sample 90 training tasks, where each task consists of 80 randomly sampled tables from the pool. Then we sample another 10 different tasks with the same procedure for the testing purpose. AutoShard is trained on the 90 training tasks. We collect the mean result on the same 10 testing tasks for all the algorithms. Note that we have purposely separated the training and testing tasks to test whether AutoShard can generalize to different table combinations from the pool.

We summarize the results in Figure~\ref{fig:compare}. We make the following observations. \textbf{First}, all the sharding algorithms outperform the random sharding, which is expected since random sharding may easily result in imbalances. \textbf{Second}, AutoShard performs significantly and consistently better than the heuristics on both synthetic and production data for both metrics, which demonstrates the effectiveness of AutoShard. \textbf{Third}, look-greedy appears to be the strongest heuristic algorithm. This is expected because it considers both table dimensions and pooling factors, which can essentially quantify the workload for the indices lookup. Nevertheless, there is still a clear gap between lookup-greedy and AutoShard. This is because AutoShard can achieve a more accurate estimation of the cost by considering indices distributions and multi-table costs, and leveraging RL to optimize the sharding process.


\subsection{Comparison with Random Search}
\label{sec:43}


To answer \textbf{Q2}, we implement a random search algorithm to understand the difficulty of identifying a strong sharding plan in the search space. We choose random search because it is shown to be a strong baseline in neural architecture search when the search space is restricted~\cite{li2020random}. Specifically, we treat the tables as decisions, whose possible choices are the device IDs, and use random search to optimize the degree of balance. We follow the setting in Section~\ref{sec:42}, which results in an extremely large search space. Note that search is infeasible in production because it requires lots of GPU resources. This experiment is designed solely for understanding of the massive search space of AutoShard.

\begin{figure}[t]
  \centering

  \begin{subfigure}[b]{0.23\textwidth}
    \centering
    \includegraphics[width=0.99\textwidth]{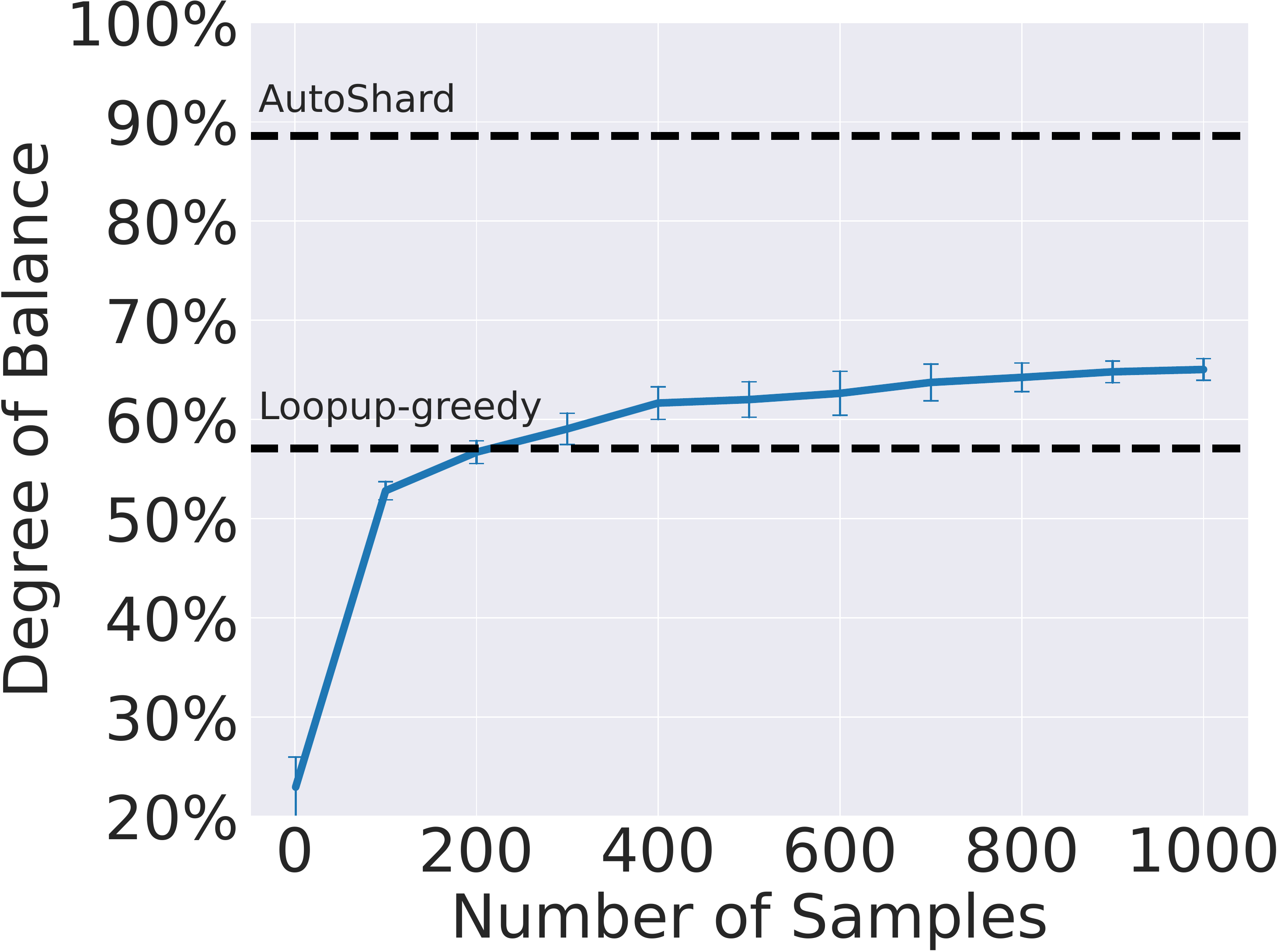}
    \caption{MetaSyn}
    \vspace{-8pt}
  \end{subfigure}%
  \begin{subfigure}[b]{0.23\textwidth}
    \centering
    \includegraphics[width=0.99\textwidth]{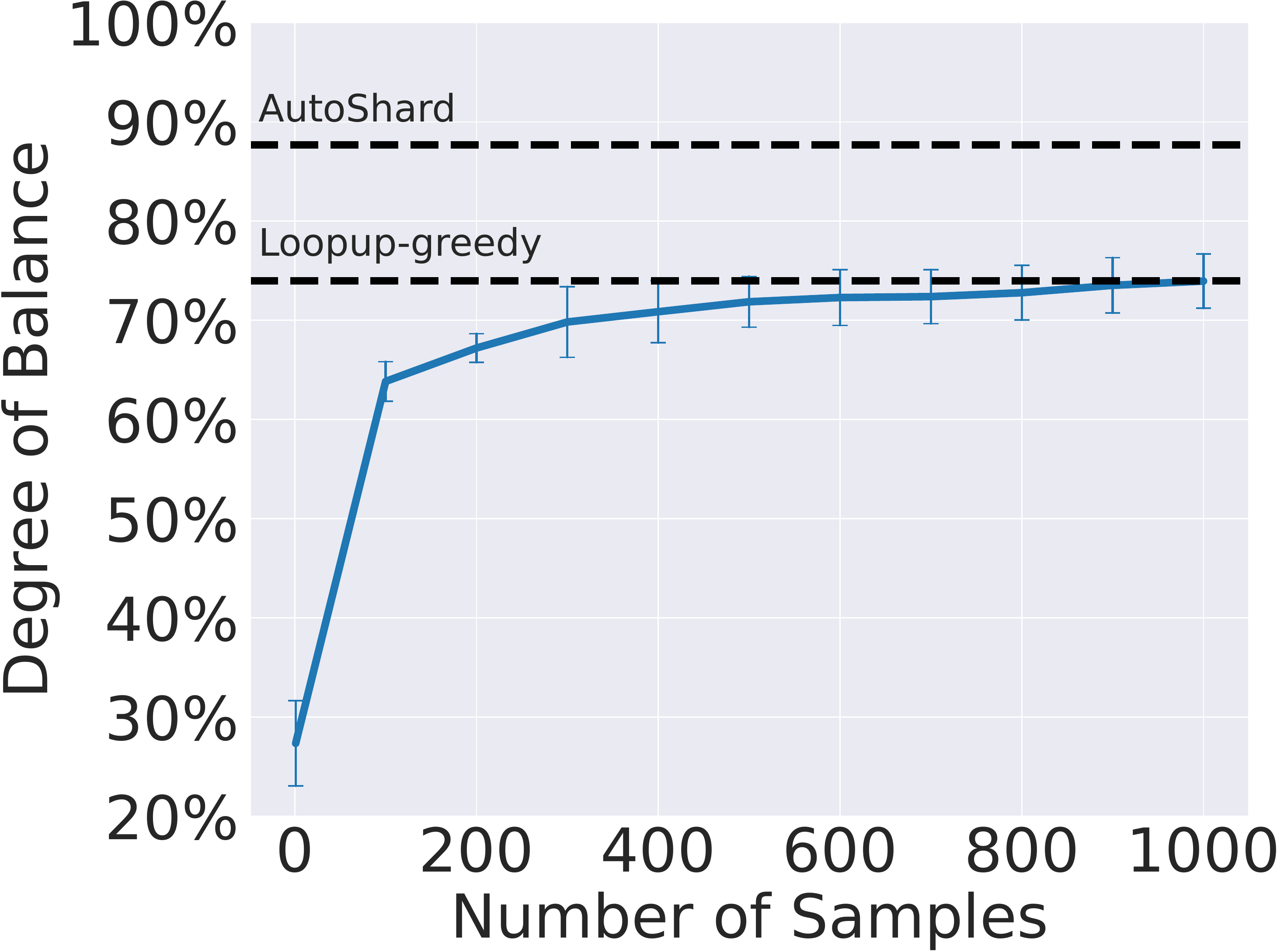}
    \caption{MetaProd}
    \vspace{-8pt}
  \end{subfigure}%
  \caption{Performance of random search with five runs. Note that searching 1000 samples is extremely time-consuming and are often impractical for production use (it takes 9,783 seconds for MetaSyn and 14,599 seconds for MetaProd). }
  \vspace{-8pt}
  \label{fig:randomsearch}
\end{figure}

Figure~\ref{fig:randomsearch} plots the performance of random search w.r.t. the number of samples. Although random search can achieve competitive performance with lookup-greedy after hours of searching, it is far behind the AutoShard, which verifies the difficulty of embedding table sharding. In contrast, AutoShard shows clear advantages in terms of both effectiveness and efficiency. It can achieve strong performance with only a forward pass without the search.

\begin{table}[]
\centering
\caption{Ablation study of AutoShard on MetaSyn.}
\vspace{-8pt}
\label{tbl:abalation}
\footnotesize
\begin{tabular}{l|c|c}
 \toprule
 & \textbf{Degree of Balance} & \textbf{Speedup} \\
 \midrule
 w/o cost modeling & 61.3\%$\pm$12.9\% & 1.420$\pm$0.203 \\
 w/o dimension feature & 87.6\%$\pm$1.5\% & 1.706$\pm$0.056 \\
 w/o hash size feature & 87.8\%$\pm$0.9\% & 1.702$\pm$0.047 \\
 w/o pooling factor feature & 45.9\%$\pm$1.9\% & 1.271$\pm$0.035 \\
 w/o size feature & 87.0\%$\pm$1.1\% & 1.683$\pm$0.049 \\
 w/o distribution features & 84.3\%$\pm$1.3\% & 1.688$\pm$0.035 \\
 \midrule
 Full version of AutoShard & \textbf{88.6\%$\pm$1.2\%} & \textbf{1.712$\pm$0.027} \\

 \bottomrule
\end{tabular}
\vspace{-10pt}
\end{table}

\subsection{Ablation Studies}
\label{sec:44}

For \textbf{Q3}, we consider several ablations: 1) we remove the cost model and only use raw features to train RL, 2) instead of sharding with RL, we greedily assign tables like the heuristics with the only difference that we use the cost model to estimate the cost, and 3) we remove each of the table features to study the feature importance.

Table~\ref{tbl:abalation} summarizes the results. \textbf{First}, we observe a significant performance drop when removing the cost model, which verifies the necessity of cost modeling. \textbf{Second}, the pooling factor feature is a very important feature, which is expected since pooling factor can indicate the number of lookup indices. \textbf{Finally}, removing either of the features will degrade the performance, which suggests the designed features are complimentary.

\begin{figure}[t]
  \centering
  \begin{subfigure}[b]{0.46\textwidth}
    \centering
    \includegraphics[width=0.99\textwidth]{figs/42_legend.pdf}
  \end{subfigure}%
 
  \begin{subfigure}[b]{0.23\textwidth}
    \centering
    \includegraphics[width=0.99\textwidth]{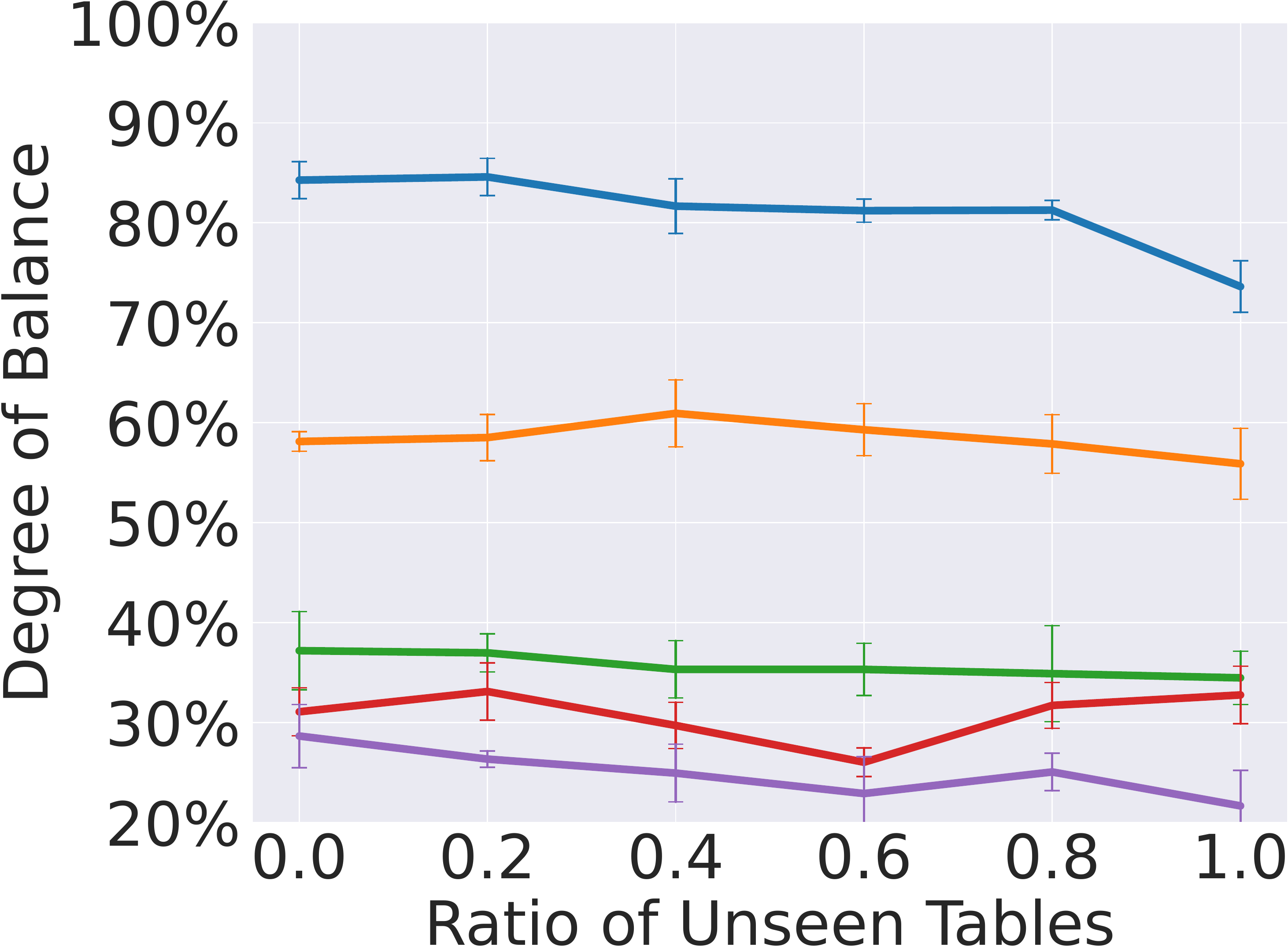}
    \vspace{-15pt}
  \end{subfigure}%
  \begin{subfigure}[b]{0.23\textwidth}
    \centering
    \includegraphics[width=0.99\textwidth]{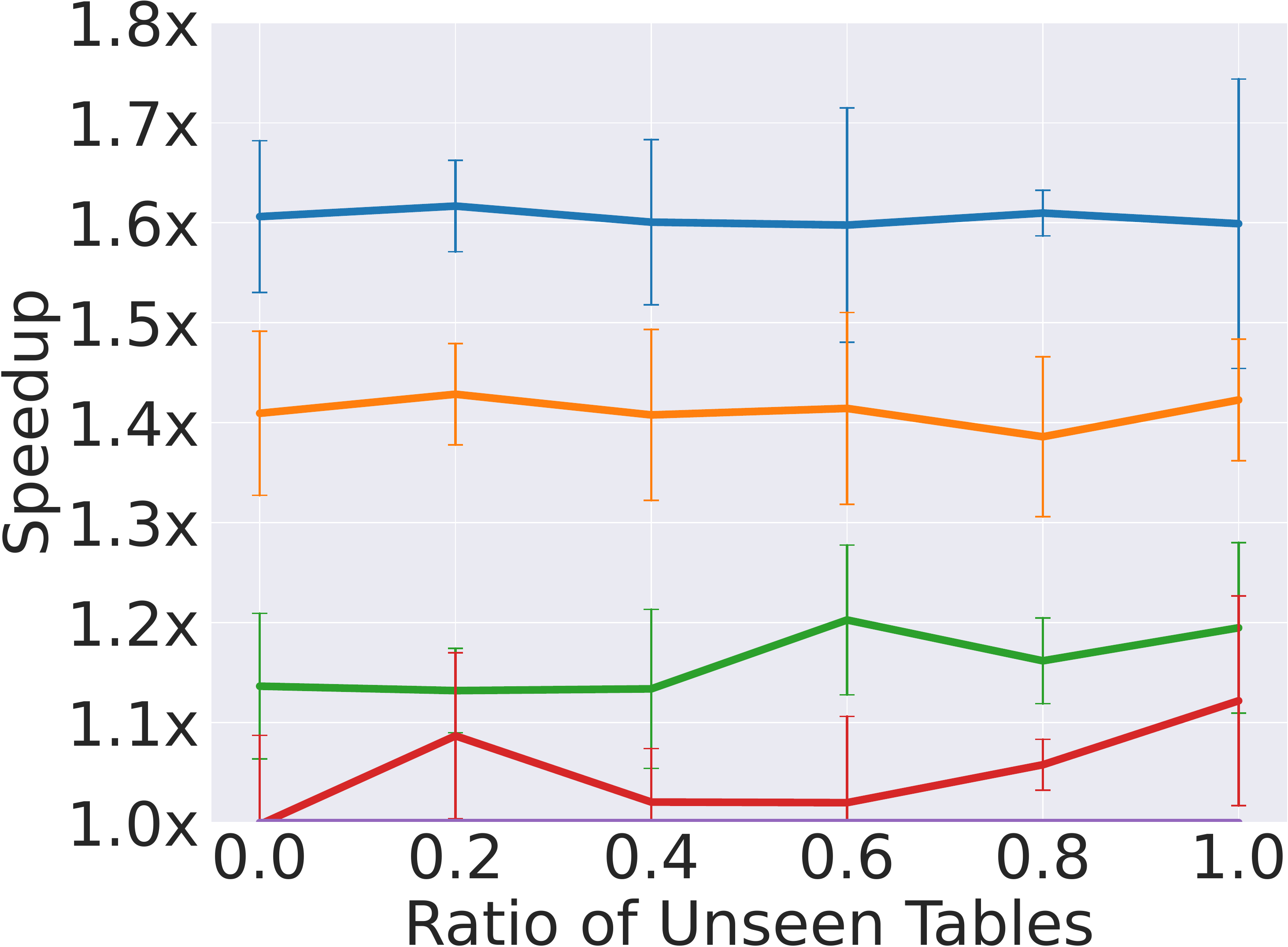}
    \vspace{-15pt}
  \end{subfigure}%
  \caption{Degree of balance (left) and speedup (right) of AutoShard on MetaSyn with different ratios of unseen tables. Note that when the ratio is 0.0, the results are worse than those shown in Figure~\ref{fig:compare}. This is because we only use half of the tables in this experiment. A possible reason is that more tables can improve the generalization ability of AutoShard, which leads to better results on the testing tasks.}
  \vspace{-10pt}
  \label{fig:unseen}
\end{figure}

\begin{figure}[t]
  \centering
  \begin{subfigure}[b]{0.46\textwidth}
    \centering
    \includegraphics[width=0.99\textwidth]{figs/42_legend.pdf}
  \end{subfigure}%
 
  \begin{subfigure}[b]{0.23\textwidth}
    \centering
    \includegraphics[width=0.99\textwidth]{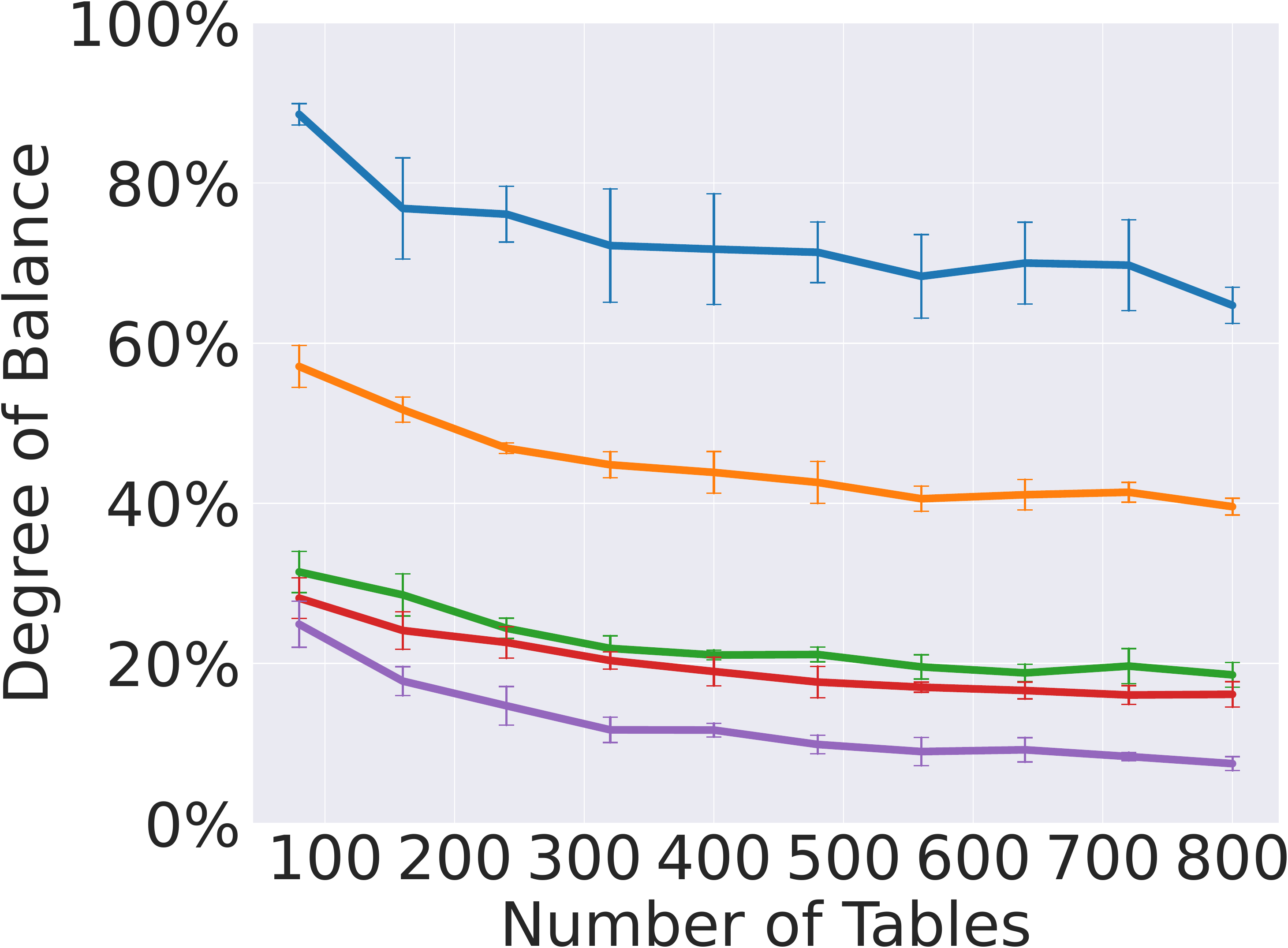}
    \vspace{-15pt}
  \end{subfigure}%
  \begin{subfigure}[b]{0.23\textwidth}
    \centering
    \includegraphics[width=0.99\textwidth]{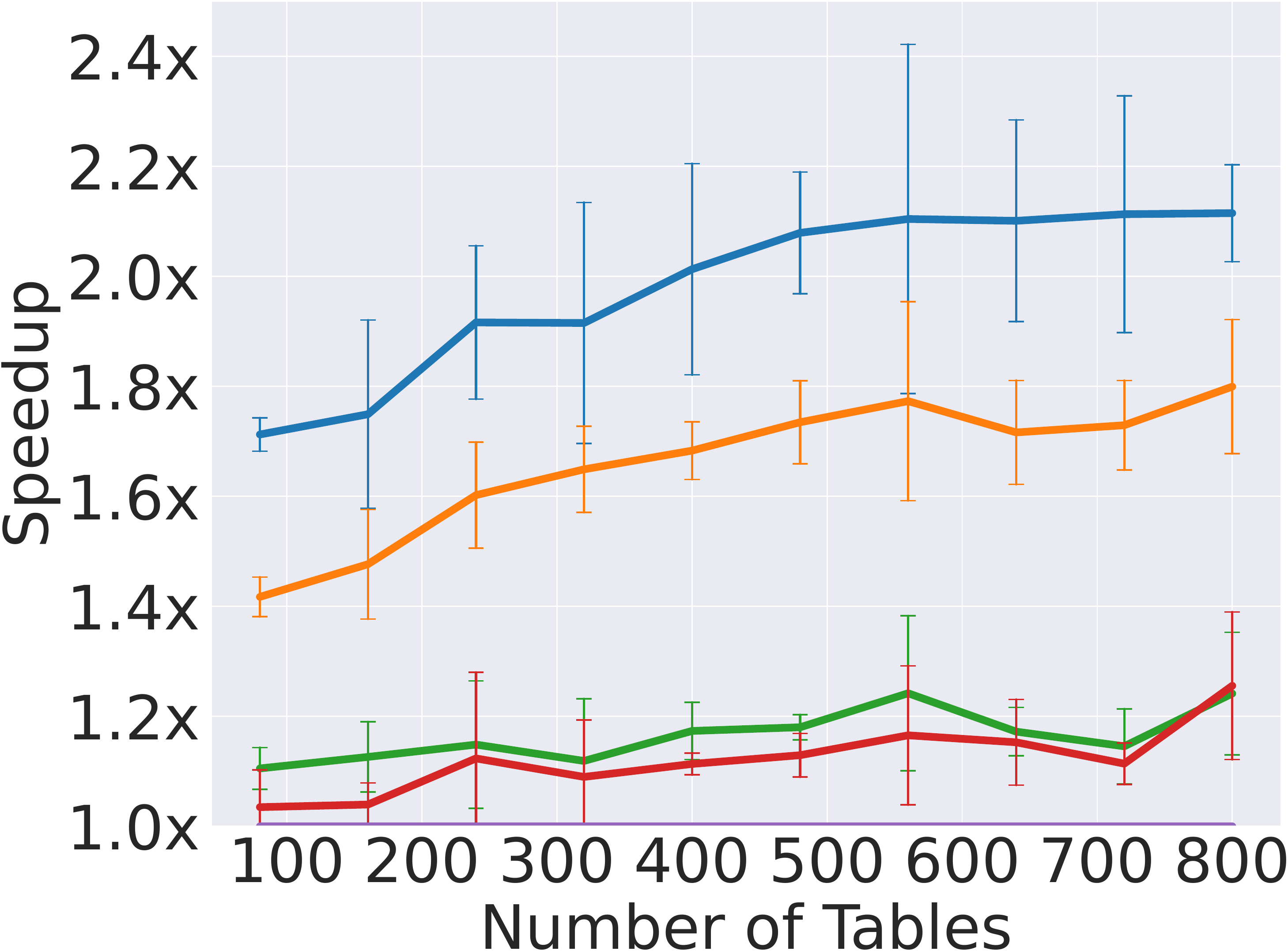}
    \vspace{-15pt}
  \end{subfigure}%
  \caption{Degree of balance (left) and speedup (right) of AutoShard on MetaSyn with up to 800 tables. We directly transfer the model trained on 80 tables without fine-tuning.}
  \vspace{-10pt}
  \label{fig:moretables}
\end{figure}

\subsection{Analysis of Transferability}
\label{sec:45}

To investigate \textbf{Q4}, we evaluate the transferability of AutoShard on unseen tables and sharding tasks with more tables.

To test AutoShard on unseen tables, we split the original table pools in half, where the first sub-pool is used to train AutoShard, and the tables in the second sub-pool are unseen in training. Then we randomly mix the tables from the two sub-pools to construct the sharding tasks based on a specified ratio of unseen tables. A ratio of 0.0 suggests that all the tables are from the first sub-pool (i.e., all the tables in the sharding tasks are seen in training), while a ratio of 1.0 means all the tables are from the second sub-pool (i.e., all the tables are unseen). A high ratio will make the transferability task more challenging. We report the performance w.r.t. different ratios of unseen tables in  Figure~\ref{fig:unseen}. \textbf{First}, AutoShard only shows a moderate performance decrease when a part of the tables are unseen. Specifically, when the ratio is between 0.0 to 0.8, AutoShard can achieve at least 80\% degree of balance and 1.6X speedup. \textbf{Second}, when all the tables are unseen (i.e., the ratio is 1.0), AutoShard can still achieve more than 70\% degree of balance and around 1.6X speedup, which significantly outperform the baselines. 

To test whether AutoShard can scale to hundreds of tables, we compare AutoShard with baselines on tasks that shard up to 800 tables to 80 GPU devices. Note that in this experiment we have not trained any new models but instead directly apply the model trained on 80 tables. We plot the results in Figure~\ref{fig:moretables} and make the following observations. \textbf{First}, the degree of balance decreases for all the sharding algorithms. This is expected because the task becomes more challenging with more tables. \textbf{Second}, AutoShard can significantly outperform the baselines in all the settings. In particular, we surprisingly observe an increase in speedup with more tables. This is because random sharding will perform poorly with more tables. This again demonstrates the superiority of AutoShard.

Overall, we conclude that AutoShard can well transfer to unseen tables and hundreds of tables, making it a desirable choice in handling complex training tasks in the production environment.


\subsection{Analysis of Training/Inference Efficiency}
\label{sec:46}

\begin{figure}[t]
  \centering

  \begin{subfigure}[b]{0.23\textwidth}
    \centering
    \includegraphics[width=0.99\textwidth]{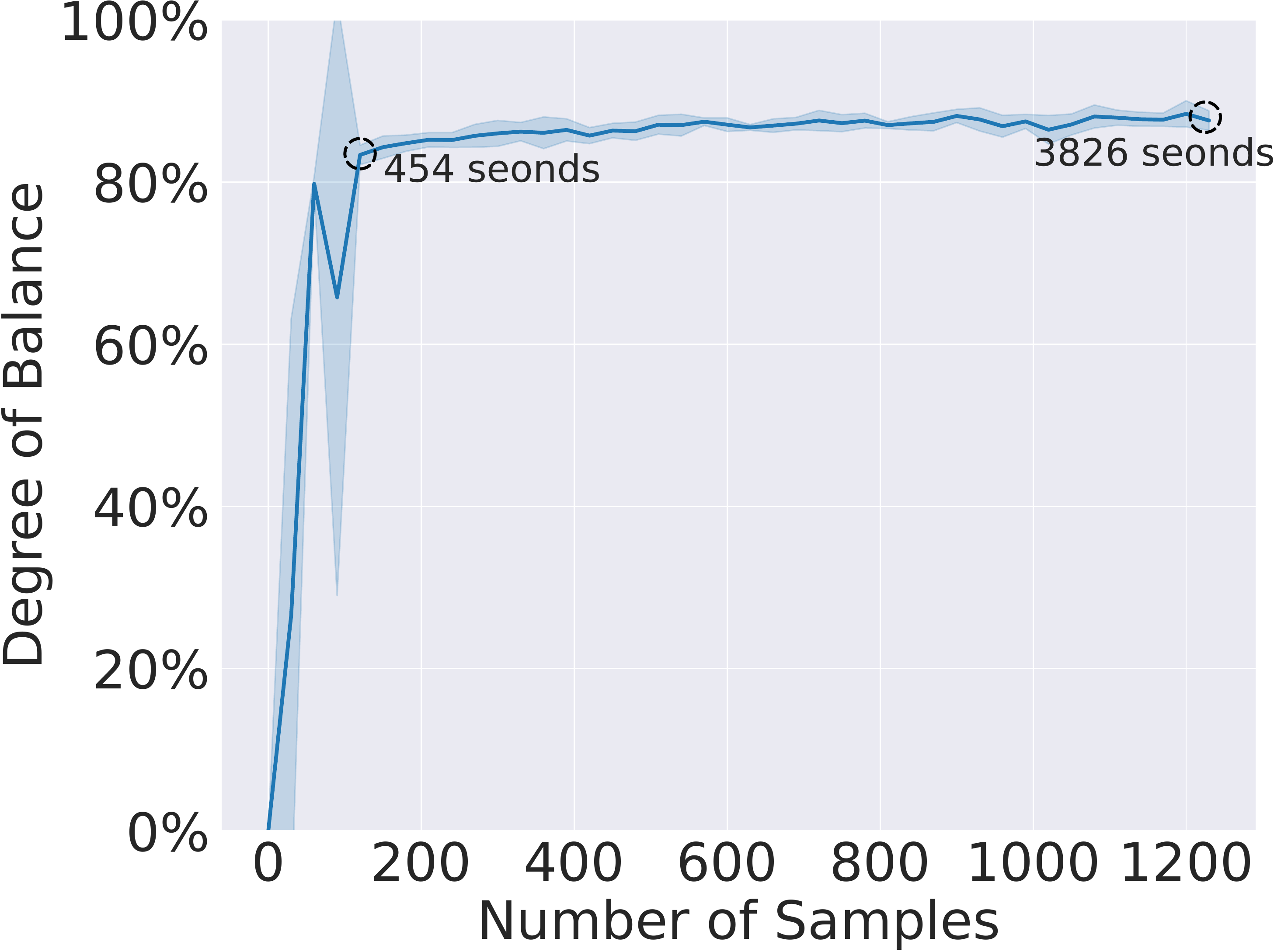}
    \vspace{-15pt}
  \end{subfigure}%
  \begin{subfigure}[b]{0.23\textwidth}
    \centering
    \includegraphics[width=0.99\textwidth]{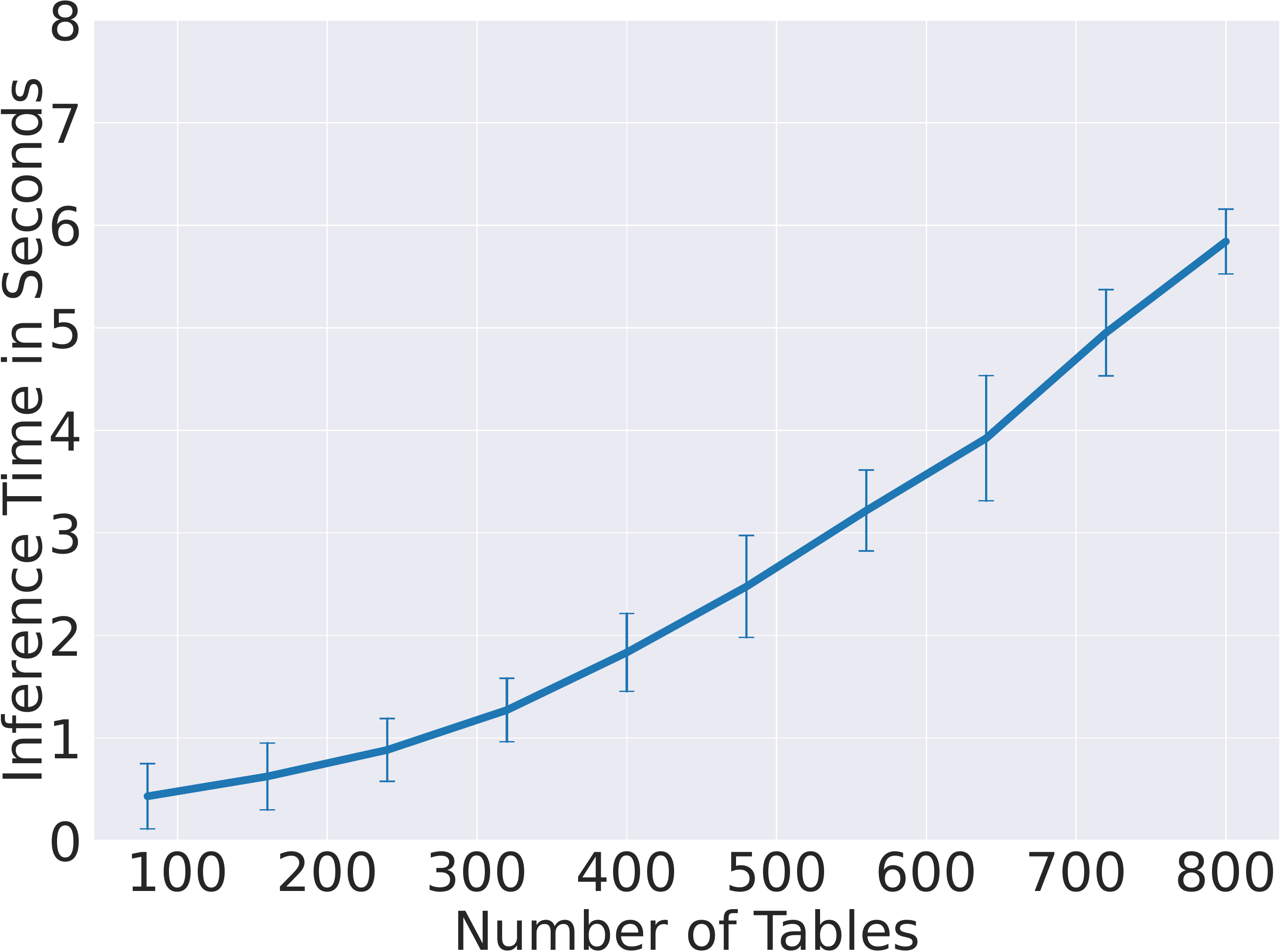}
    \vspace{-15pt}
  \end{subfigure}%
  \caption{Training curve on four 2080 Ti GPUs (left) and inference time with a single CPU core (right). }
  \vspace{-20pt}
  \label{fig:efficiency}
\end{figure}

We analyze the training and inference time to answer \textbf{Q5}. The left-hand side of Figure~\ref{fig:efficiency} plots the training curve of AutoShard on four GPUs. We observe that AutoShard can achieve more than 80\% degree of balance within around 150 samples or 454 seconds. This is highly efficient for production use since we only need to train AutoShard offline periodically (e.g. we can run a daily or weekly training job). The right-hand side of Figure~\ref{fig:efficiency} shows the inference time with a single CPU core. AutoShard can shard hundreds of tables in seconds. This cost is neglectable in production use.

\section{Related Work}

\textbf{Deep recommendation models.} Deep-learning-based recommendation models have shown superior performance in many recommendation scenarios~\cite{zhang2019deep,cheng2016wide,naumov2019deep,he2017neural}. Due to the ultra-large-scale of the data and the features in industrial applications, distributed training solutions have been developed to improve the training efficiency~\cite{acun2021understanding,covington2016deep,zhou2019deep,liu2017related,gomez2015netflix}. Despite these efforts, embedding table sharding remains to be a critical challenge for distributed training. AutoShard is the first learning-based sharding algorithm that can optimize the sharding strategy in an end-to-end fashion.


\textbf{Tackling large embedding tables.} How to deal with the ultra-large embedding tables of recommendation models has been a long-standing challenge. One line of work aims to reduce the embedding table size, such as sharing the embeddings across related features~\cite{zhang2020model,shi2020compositional}, searching the vocabulary sizes or the table dimensions~\cite{zhao2020autoemb,joglekar2020neural}, pruning~\cite{liu2020learnable}, quantilization~\cite{kang2020learning}, and hashing~\cite{kang2021learning}. Our work is orthogonal to these methods since AutoShard can be applied to compressed tables as well. A related work explored using the tiered memory hierarchy to store the embedding tables~\cite{sethi2022recshard}. They exploited the unequal access patterns of embedding tables to improve the efficiency by placing hot rows in the GPU memory. Our efforts complement~\cite{sethi2022recshard} by providing an end-to-end learning-based framework for cost approximation and partitioning optimization.



\textbf{Deep RL.} Deep RL has shown promise in accomplishing goal-oriented tasks~\cite{mnih2013playing,silver2017mastering,zha2021douzero,zha2020rank,zha2021rlcard}. Recently, deep RL has been applied to various machine learning model design tasks, such as neural architecture search~\cite{zoph2016neural}, pipeline search~\cite{zha2021autovideo,lai2021tods,li2021automated}, data augmentation/sampling~\cite{cubuk2018autoaugment,zha2019experience,zha2020meta}. Our work also falls into this line of studies but we instead focus on optimizing the model efficiency. Our work is also related to applying RL for classical combinational optimization~\cite{barrett2020exploratory}. Unlike~\cite{barrett2020exploratory}, we tackle a real-world combinational optimization challenge in industrial recommender systems.

\section{Conclusions and Future Work}
This work presents a novel solution for embedding table sharding practiced at Meta. The proposed algorithm, namely AutoShard, uses a cost model to efficiently estimate the table cost and leverages deep RL to solve the partition problem. The empirical results suggest that AutoShard is effective, transferable, and efficient. Through developing AutoShard, we show the promise in applying RL to optimize the industrial-level system designs. We have open-sourced a prototype of AutoShard to motivate and facilitate future exploration in this direction. In the future, we will extend AutoShard to tackle more complex sharding tasks by modeling the cost of the communication and the tiered memory hierarchy.

\section*{Acknowledgements}
The work is, in part, supported by NSF (\#IIS-1900990, \#IIS-1750074). The views and conclusions in this paper are those of the authors and should not be interpreted as representing any funding agencies.



\bibliographystyle{ACM-Reference-Format}
\bibliography{ref}


\begin{thebibliography}{41}


\ifx \showCODEN    \undefined \def \showCODEN     #1{\unskip}     \fi
\ifx \showDOI      \undefined \def \showDOI       #1{#1}\fi
\ifx \showISBNx    \undefined \def \showISBNx     #1{\unskip}     \fi
\ifx \showISBNxiii \undefined \def \showISBNxiii  #1{\unskip}     \fi
\ifx \showISSN     \undefined \def \showISSN      #1{\unskip}     \fi
\ifx \showLCCN     \undefined \def \showLCCN      #1{\unskip}     \fi
\ifx \shownote     \undefined \def \shownote      #1{#1}          \fi
\ifx \showarticletitle \undefined \def \showarticletitle #1{#1}   \fi
\ifx \showURL      \undefined \def \showURL       {\relax}        \fi
\providecommand\bibfield[2]{#2}
\providecommand\bibinfo[2]{#2}
\providecommand\natexlab[1]{#1}
\providecommand\showeprint[2][]{arXiv:#2}

\bibitem[\protect\citeauthoryear{??}{ama}{[n.d.]}]%
        {amazon-dsstne}
 \bibinfo{year}{[n.d.]}\natexlab{}.
\newblock \bibinfo{title}{Amazon DSSTNE: Deep Scalable Sparse Tensor Network
  Engine}.
\newblock
  \bibinfo{howpublished}{\url{https://github.com/amazon-archives/amazon-dsstne}}.
\newblock


\bibitem[\protect\citeauthoryear{Acun, Murphy, Wang, Nie, Wu, and
  Hazelwood}{Acun et~al\mbox{.}}{2021}]%
        {acun2021understanding}
\bibfield{author}{\bibinfo{person}{Bilge Acun}, \bibinfo{person}{Matthew
  Murphy}, \bibinfo{person}{Xiaodong Wang}, \bibinfo{person}{Jade Nie},
  \bibinfo{person}{Carole-Jean Wu}, {and} \bibinfo{person}{Kim Hazelwood}.}
  \bibinfo{year}{2021}\natexlab{}.
\newblock \showarticletitle{Understanding training efficiency of deep learning
  recommendation models at scale}. In \bibinfo{booktitle}{\emph{HPCA}}.
\newblock


\bibitem[\protect\citeauthoryear{Barrett, Clements, Foerster, and
  Lvovsky}{Barrett et~al\mbox{.}}{2020}]%
        {barrett2020exploratory}
\bibfield{author}{\bibinfo{person}{Thomas Barrett}, \bibinfo{person}{William
  Clements}, \bibinfo{person}{Jakob Foerster}, {and} \bibinfo{person}{Alex
  Lvovsky}.} \bibinfo{year}{2020}\natexlab{}.
\newblock \showarticletitle{Exploratory combinatorial optimization with
  reinforcement learning}. In \bibinfo{booktitle}{\emph{AAAI}}.
\newblock


\bibitem[\protect\citeauthoryear{Cheng, Koc, Harmsen, Shaked, Chandra, Aradhye,
  Anderson, Corrado, Chai, Ispir, et~al\mbox{.}}{Cheng et~al\mbox{.}}{2016}]%
        {cheng2016wide}
\bibfield{author}{\bibinfo{person}{Heng-Tze Cheng}, \bibinfo{person}{Levent
  Koc}, \bibinfo{person}{Jeremiah Harmsen}, \bibinfo{person}{Tal Shaked},
  \bibinfo{person}{Tushar Chandra}, \bibinfo{person}{Hrishi Aradhye},
  \bibinfo{person}{Glen Anderson}, \bibinfo{person}{Greg Corrado},
  \bibinfo{person}{Wei Chai}, \bibinfo{person}{Mustafa Ispir}, {et~al\mbox{.}}}
  \bibinfo{year}{2016}\natexlab{}.
\newblock \showarticletitle{Wide \& deep learning for recommender systems}. In
  \bibinfo{booktitle}{\emph{DLRS Workshop}}.
\newblock


\bibitem[\protect\citeauthoryear{Covington, Adams, and Sargin}{Covington
  et~al\mbox{.}}{2016}]%
        {covington2016deep}
\bibfield{author}{\bibinfo{person}{Paul Covington}, \bibinfo{person}{Jay
  Adams}, {and} \bibinfo{person}{Emre Sargin}.}
  \bibinfo{year}{2016}\natexlab{}.
\newblock \showarticletitle{Deep neural networks for youtube recommendations}.
  In \bibinfo{booktitle}{\emph{RecSys}}.
\newblock


\bibitem[\protect\citeauthoryear{Cubuk, Zoph, Mane, Vasudevan, and Le}{Cubuk
  et~al\mbox{.}}{2019}]%
        {cubuk2018autoaugment}
\bibfield{author}{\bibinfo{person}{Ekin~D Cubuk}, \bibinfo{person}{Barret
  Zoph}, \bibinfo{person}{Dandelion Mane}, \bibinfo{person}{Vijay Vasudevan},
  {and} \bibinfo{person}{Quoc~V Le}.} \bibinfo{year}{2019}\natexlab{}.
\newblock \showarticletitle{Autoaugment: Learning augmentation policies from
  data}. In \bibinfo{booktitle}{\emph{CVPR}}.
\newblock


\bibitem[\protect\citeauthoryear{Espeholt, Soyer, Munos, Simonyan, Mnih, Ward,
  Doron, Firoiu, Harley, et~al\mbox{.}}{Espeholt et~al\mbox{.}}{2018}]%
        {espeholt2018impala}
\bibfield{author}{\bibinfo{person}{Lasse Espeholt}, \bibinfo{person}{Hubert
  Soyer}, \bibinfo{person}{Remi Munos}, \bibinfo{person}{Karen Simonyan},
  \bibinfo{person}{Vlad Mnih}, \bibinfo{person}{Tom Ward},
  \bibinfo{person}{Yotam Doron}, \bibinfo{person}{Vlad Firoiu},
  \bibinfo{person}{Tim Harley}, {et~al\mbox{.}}}
  \bibinfo{year}{2018}\natexlab{}.
\newblock \showarticletitle{Impala: Scalable distributed deep-rl with
  importance weighted actor-learner architectures}. In
  \bibinfo{booktitle}{\emph{ICML}}.
\newblock


\bibitem[\protect\citeauthoryear{Gomez-Uribe and Hunt}{Gomez-Uribe and
  Hunt}{2015}]%
        {gomez2015netflix}
\bibfield{author}{\bibinfo{person}{Carlos~A Gomez-Uribe} {and}
  \bibinfo{person}{Neil Hunt}.} \bibinfo{year}{2015}\natexlab{}.
\newblock \showarticletitle{The netflix recommender system: Algorithms,
  business value, and innovation}.
\newblock \bibinfo{journal}{\emph{ACM Transactions on Management Information
  Systems (TMIS)}} \bibinfo{volume}{6}, \bibinfo{number}{4}
  (\bibinfo{year}{2015}), \bibinfo{pages}{1--19}.
\newblock


\bibitem[\protect\citeauthoryear{Gupta, Wu, Wang, Naumov, Reagen, Brooks,
  Cottel, Hazelwood, Hempstead, Jia, et~al\mbox{.}}{Gupta
  et~al\mbox{.}}{2020}]%
        {gupta2020architectural}
\bibfield{author}{\bibinfo{person}{Udit Gupta}, \bibinfo{person}{Carole-Jean
  Wu}, \bibinfo{person}{Xiaodong Wang}, \bibinfo{person}{Maxim Naumov},
  \bibinfo{person}{Brandon Reagen}, \bibinfo{person}{David Brooks},
  \bibinfo{person}{Bradford Cottel}, \bibinfo{person}{Kim Hazelwood},
  \bibinfo{person}{Mark Hempstead}, \bibinfo{person}{Bill Jia},
  {et~al\mbox{.}}} \bibinfo{year}{2020}\natexlab{}.
\newblock \showarticletitle{The architectural implications of facebook's
  dnn-based personalized recommendation}. In \bibinfo{booktitle}{\emph{HPCA}}.
\newblock


\bibitem[\protect\citeauthoryear{He, Liao, Zhang, Nie, Hu, and Chua}{He
  et~al\mbox{.}}{2017}]%
        {he2017neural}
\bibfield{author}{\bibinfo{person}{Xiangnan He}, \bibinfo{person}{Lizi Liao},
  \bibinfo{person}{Hanwang Zhang}, \bibinfo{person}{Liqiang Nie},
  \bibinfo{person}{Xia Hu}, {and} \bibinfo{person}{Tat-Seng Chua}.}
  \bibinfo{year}{2017}\natexlab{}.
\newblock \showarticletitle{Neural collaborative filtering}. In
  \bibinfo{booktitle}{\emph{WWW}}.
\newblock


\bibitem[\protect\citeauthoryear{Jeong and Hwang}{Jeong and Hwang}{2018}]%
        {jeong2018nonvolatile}
\bibfield{author}{\bibinfo{person}{Doo~Seok Jeong} {and}
  \bibinfo{person}{Cheol~Seong Hwang}.} \bibinfo{year}{2018}\natexlab{}.
\newblock \showarticletitle{Nonvolatile memory materials for neuromorphic
  intelligent machines}.
\newblock \bibinfo{journal}{\emph{Advanced Materials}} \bibinfo{volume}{30},
  \bibinfo{number}{42} (\bibinfo{year}{2018}), \bibinfo{pages}{1704729}.
\newblock


\bibitem[\protect\citeauthoryear{Jiang, Deng, Yi, Hu, Zhou, Zheng, Huang, Guo,
  Wang, Song, et~al\mbox{.}}{Jiang et~al\mbox{.}}{2019}]%
        {jiang2019xdl}
\bibfield{author}{\bibinfo{person}{Biye Jiang}, \bibinfo{person}{Chao Deng},
  \bibinfo{person}{Huimin Yi}, \bibinfo{person}{Zelin Hu},
  \bibinfo{person}{Guorui Zhou}, \bibinfo{person}{Yang Zheng},
  \bibinfo{person}{Sui Huang}, \bibinfo{person}{Xinyang Guo},
  \bibinfo{person}{Dongyue Wang}, \bibinfo{person}{Yue Song}, {et~al\mbox{.}}}
  \bibinfo{year}{2019}\natexlab{}.
\newblock \showarticletitle{XDL: an industrial deep learning framework for
  high-dimensional sparse data}. In \bibinfo{booktitle}{\emph{KDD Workshop}}.
\newblock


\bibitem[\protect\citeauthoryear{Joglekar, Li, Chen, Xu, Wang, Adams, Khaitan,
  Liu, and Le}{Joglekar et~al\mbox{.}}{2020}]%
        {joglekar2020neural}
\bibfield{author}{\bibinfo{person}{Manas~R Joglekar}, \bibinfo{person}{Cong
  Li}, \bibinfo{person}{Mei Chen}, \bibinfo{person}{Taibai Xu},
  \bibinfo{person}{Xiaoming Wang}, \bibinfo{person}{Jay~K Adams},
  \bibinfo{person}{Pranav Khaitan}, \bibinfo{person}{Jiahui Liu}, {and}
  \bibinfo{person}{Quoc~V Le}.} \bibinfo{year}{2020}\natexlab{}.
\newblock \showarticletitle{Neural input search for large scale recommendation
  models}. In \bibinfo{booktitle}{\emph{KDD}}.
\newblock


\bibitem[\protect\citeauthoryear{Kang, Cheng, Chen, Yi, Lin, Hong, and
  Chi}{Kang et~al\mbox{.}}{2020}]%
        {kang2020learning}
\bibfield{author}{\bibinfo{person}{Wang-Cheng Kang},
  \bibinfo{person}{Derek~Zhiyuan Cheng}, \bibinfo{person}{Ting Chen},
  \bibinfo{person}{Xinyang Yi}, \bibinfo{person}{Dong Lin},
  \bibinfo{person}{Lichan Hong}, {and} \bibinfo{person}{Ed~H Chi}.}
  \bibinfo{year}{2020}\natexlab{}.
\newblock \showarticletitle{Learning multi-granular quantized embeddings for
  large-vocab categorical features in recommender systems}. In
  \bibinfo{booktitle}{\emph{WWW}}.
\newblock


\bibitem[\protect\citeauthoryear{Kang, Cheng, Yao, Yi, Chen, Hong, and
  Chi}{Kang et~al\mbox{.}}{2021}]%
        {kang2021learning}
\bibfield{author}{\bibinfo{person}{Wang-Cheng Kang},
  \bibinfo{person}{Derek~Zhiyuan Cheng}, \bibinfo{person}{Tiansheng Yao},
  \bibinfo{person}{Xinyang Yi}, \bibinfo{person}{Ting Chen},
  \bibinfo{person}{Lichan Hong}, {and} \bibinfo{person}{Ed~H Chi}.}
  \bibinfo{year}{2021}\natexlab{}.
\newblock \showarticletitle{Learning to Embed Categorical Features without
  Embedding Tables for Recommendation}. In \bibinfo{booktitle}{\emph{KDD}}.
\newblock


\bibitem[\protect\citeauthoryear{Khudia, Huang, Basu, Deng, Liu, Park, and
  Smelyanskiy}{Khudia et~al\mbox{.}}{2021}]%
        {fbgemm}
\bibfield{author}{\bibinfo{person}{Daya Khudia}, \bibinfo{person}{Jianyu
  Huang}, \bibinfo{person}{Protonu Basu}, \bibinfo{person}{Summer Deng},
  \bibinfo{person}{Haixin Liu}, \bibinfo{person}{Jongsoo Park}, {and}
  \bibinfo{person}{Mikhail Smelyanskiy}.} \bibinfo{year}{2021}\natexlab{}.
\newblock \showarticletitle{FBGEMM: Enabling High-Performance Low-Precision
  Deep Learning Inference}.
\newblock \bibinfo{journal}{\emph{arXiv preprint arXiv:2101.05615}}
  (\bibinfo{year}{2021}).
\newblock


\bibitem[\protect\citeauthoryear{K{\"u}ttler, Nardelli, Lavril, Selvatici,
  Sivakumar, Rockt{\"a}schel, and Grefenstette}{K{\"u}ttler
  et~al\mbox{.}}{2019}]%
        {kuttler2019torchbeast}
\bibfield{author}{\bibinfo{person}{Heinrich K{\"u}ttler},
  \bibinfo{person}{Nantas Nardelli}, \bibinfo{person}{Thibaut Lavril},
  \bibinfo{person}{Marco Selvatici}, \bibinfo{person}{Viswanath Sivakumar},
  \bibinfo{person}{Tim Rockt{\"a}schel}, {and} \bibinfo{person}{Edward
  Grefenstette}.} \bibinfo{year}{2019}\natexlab{}.
\newblock \showarticletitle{Torchbeast: A pytorch platform for distributed rl}.
\newblock \bibinfo{journal}{\emph{arXiv preprint arXiv:1910.03552}}
  (\bibinfo{year}{2019}).
\newblock


\bibitem[\protect\citeauthoryear{Lai, Zha, Wang, Xu, Zhao, Kumar, Chen,
  Zumkhawaka, Wan, Martinez, et~al\mbox{.}}{Lai et~al\mbox{.}}{2021}]%
        {lai2021tods}
\bibfield{author}{\bibinfo{person}{Kwei-Herng Lai}, \bibinfo{person}{Daochen
  Zha}, \bibinfo{person}{Guanchu Wang}, \bibinfo{person}{Junjie Xu},
  \bibinfo{person}{Yue Zhao}, \bibinfo{person}{Devesh Kumar},
  \bibinfo{person}{Yile Chen}, \bibinfo{person}{Purav Zumkhawaka},
  \bibinfo{person}{Minyang Wan}, \bibinfo{person}{Diego Martinez},
  {et~al\mbox{.}}} \bibinfo{year}{2021}\natexlab{}.
\newblock \showarticletitle{TODS: An Automated Time Series Outlier Detection
  System}. In \bibinfo{booktitle}{\emph{AAAI}}.
\newblock


\bibitem[\protect\citeauthoryear{Li and Talwalkar}{Li and Talwalkar}{2020}]%
        {li2020random}
\bibfield{author}{\bibinfo{person}{Liam Li} {and} \bibinfo{person}{Ameet
  Talwalkar}.} \bibinfo{year}{2020}\natexlab{}.
\newblock \showarticletitle{Random search and reproducibility for neural
  architecture search}. In \bibinfo{booktitle}{\emph{UAI}}.
\newblock


\bibitem[\protect\citeauthoryear{Li, Chen, Zha, Zhou, Jin, Chen, and Hu}{Li
  et~al\mbox{.}}{2021}]%
        {li2021automated}
\bibfield{author}{\bibinfo{person}{Yuening Li}, \bibinfo{person}{Zhengzhang
  Chen}, \bibinfo{person}{Daochen Zha}, \bibinfo{person}{Kaixiong Zhou},
  \bibinfo{person}{Haifeng Jin}, \bibinfo{person}{Haifeng Chen}, {and}
  \bibinfo{person}{Xia Hu}.} \bibinfo{year}{2021}\natexlab{}.
\newblock \showarticletitle{Automated Anomaly Detection via Curiosity-Guided
  Search and Self-Imitation Learning}.
\newblock \bibinfo{journal}{\emph{IEEE Transactions on Neural Networks and
  Learning Systems}} (\bibinfo{year}{2021}).
\newblock


\bibitem[\protect\citeauthoryear{Liu, Rogers, Shiau, Kislyuk, Ma, Zhong, Liu,
  and Jing}{Liu et~al\mbox{.}}{2017}]%
        {liu2017related}
\bibfield{author}{\bibinfo{person}{David~C Liu}, \bibinfo{person}{Stephanie
  Rogers}, \bibinfo{person}{Raymond Shiau}, \bibinfo{person}{Dmitry Kislyuk},
  \bibinfo{person}{Kevin~C Ma}, \bibinfo{person}{Zhigang Zhong},
  \bibinfo{person}{Jenny Liu}, {and} \bibinfo{person}{Yushi Jing}.}
  \bibinfo{year}{2017}\natexlab{}.
\newblock \showarticletitle{Related pins at pinterest: The evolution of a
  real-world recommender system}. In \bibinfo{booktitle}{\emph{WWW}}.
\newblock


\bibitem[\protect\citeauthoryear{Liu, Gao, Chen, Jin, and Li}{Liu
  et~al\mbox{.}}{2021}]%
        {liu2020learnable}
\bibfield{author}{\bibinfo{person}{Siyi Liu}, \bibinfo{person}{Chen Gao},
  \bibinfo{person}{Yihong Chen}, \bibinfo{person}{Depeng Jin}, {and}
  \bibinfo{person}{Yong Li}.} \bibinfo{year}{2021}\natexlab{}.
\newblock \showarticletitle{Learnable Embedding sizes for Recommender Systems}.
  In \bibinfo{booktitle}{\emph{ICLR}}.
\newblock


\bibitem[\protect\citeauthoryear{Mnih, Kavukcuoglu, Silver, Graves, Antonoglou,
  Wierstra, and Riedmiller}{Mnih et~al\mbox{.}}{2013}]%
        {mnih2013playing}
\bibfield{author}{\bibinfo{person}{Volodymyr Mnih}, \bibinfo{person}{Koray
  Kavukcuoglu}, \bibinfo{person}{David Silver}, \bibinfo{person}{Alex Graves},
  \bibinfo{person}{Ioannis Antonoglou}, \bibinfo{person}{Daan Wierstra}, {and}
  \bibinfo{person}{Martin Riedmiller}.} \bibinfo{year}{2013}\natexlab{}.
\newblock \showarticletitle{Playing atari with deep reinforcement learning}.
\newblock \bibinfo{journal}{\emph{arXiv preprint arXiv:1312.5602}}
  (\bibinfo{year}{2013}).
\newblock


\bibitem[\protect\citeauthoryear{Naumov, Kim, Mudigere, Sridharan, Wang, Zhao,
  Yilmaz, Kim, Yuen, Ozdal, et~al\mbox{.}}{Naumov et~al\mbox{.}}{2020}]%
        {naumov2020deep}
\bibfield{author}{\bibinfo{person}{Maxim Naumov}, \bibinfo{person}{John Kim},
  \bibinfo{person}{Dheevatsa Mudigere}, \bibinfo{person}{Srinivas Sridharan},
  \bibinfo{person}{Xiaodong Wang}, \bibinfo{person}{Whitney Zhao},
  \bibinfo{person}{Serhat Yilmaz}, \bibinfo{person}{Changkyu Kim},
  \bibinfo{person}{Hector Yuen}, \bibinfo{person}{Mustafa Ozdal},
  {et~al\mbox{.}}} \bibinfo{year}{2020}\natexlab{}.
\newblock \showarticletitle{Deep learning training in facebook data centers:
  Design of scale-up and scale-out systems}.
\newblock \bibinfo{journal}{\emph{arXiv preprint arXiv:2003.09518}}
  (\bibinfo{year}{2020}).
\newblock


\bibitem[\protect\citeauthoryear{Naumov, Mudigere, Shi, Huang, Sundaraman,
  Park, Wang, Gupta, Wu, Azzolini, et~al\mbox{.}}{Naumov et~al\mbox{.}}{2019}]%
        {naumov2019deep}
\bibfield{author}{\bibinfo{person}{Maxim Naumov}, \bibinfo{person}{Dheevatsa
  Mudigere}, \bibinfo{person}{Hao-Jun~Michael Shi}, \bibinfo{person}{Jianyu
  Huang}, \bibinfo{person}{Narayanan Sundaraman}, \bibinfo{person}{Jongsoo
  Park}, \bibinfo{person}{Xiaodong Wang}, \bibinfo{person}{Udit Gupta},
  \bibinfo{person}{Carole-Jean Wu}, \bibinfo{person}{Alisson~G Azzolini},
  {et~al\mbox{.}}} \bibinfo{year}{2019}\natexlab{}.
\newblock \showarticletitle{Deep learning recommendation model for
  personalization and recommendation systems}.
\newblock \bibinfo{journal}{\emph{arXiv preprint arXiv:1906.00091}}
  (\bibinfo{year}{2019}).
\newblock


\bibitem[\protect\citeauthoryear{Sethi, Acun, Agarwal, Kozyrakis, Trippel, and
  Wu}{Sethi et~al\mbox{.}}{2022}]%
        {sethi2022recshard}
\bibfield{author}{\bibinfo{person}{Geet Sethi}, \bibinfo{person}{Bilge Acun},
  \bibinfo{person}{Niket Agarwal}, \bibinfo{person}{Christos Kozyrakis},
  \bibinfo{person}{Caroline Trippel}, {and} \bibinfo{person}{Carole-Jean Wu}.}
  \bibinfo{year}{2022}\natexlab{}.
\newblock \showarticletitle{RecShard: Statistical Feature-Based Memory
  Optimization for Industry-Scale Neural Recommendation}. In
  \bibinfo{booktitle}{\emph{ASPLOS}}.
\newblock


\bibitem[\protect\citeauthoryear{Shi, Mudigere, Naumov, and Yang}{Shi
  et~al\mbox{.}}{2020}]%
        {shi2020compositional}
\bibfield{author}{\bibinfo{person}{Hao-Jun~Michael Shi},
  \bibinfo{person}{Dheevatsa Mudigere}, \bibinfo{person}{Maxim Naumov}, {and}
  \bibinfo{person}{Jiyan Yang}.} \bibinfo{year}{2020}\natexlab{}.
\newblock \showarticletitle{Compositional embeddings using complementary
  partitions for memory-efficient recommendation systems}. In
  \bibinfo{booktitle}{\emph{KDD}}.
\newblock


\bibitem[\protect\citeauthoryear{Silver, Schrittwieser, Simonyan, Antonoglou,
  Huang, Guez, Hubert, Baker, Lai, Bolton, et~al\mbox{.}}{Silver
  et~al\mbox{.}}{2017}]%
        {silver2017mastering}
\bibfield{author}{\bibinfo{person}{David Silver}, \bibinfo{person}{Julian
  Schrittwieser}, \bibinfo{person}{Karen Simonyan}, \bibinfo{person}{Ioannis
  Antonoglou}, \bibinfo{person}{Aja Huang}, \bibinfo{person}{Arthur Guez},
  \bibinfo{person}{Thomas Hubert}, \bibinfo{person}{Lucas Baker},
  \bibinfo{person}{Matthew Lai}, \bibinfo{person}{Adrian Bolton},
  {et~al\mbox{.}}} \bibinfo{year}{2017}\natexlab{}.
\newblock \showarticletitle{Mastering the game of go without human knowledge}.
\newblock \bibinfo{journal}{\emph{nature}} (\bibinfo{year}{2017}).
\newblock


\bibitem[\protect\citeauthoryear{Song, Shi, Xiao, Duan, Xu, Zhang, and
  Tang}{Song et~al\mbox{.}}{2019}]%
        {song2019autoint}
\bibfield{author}{\bibinfo{person}{Weiping Song}, \bibinfo{person}{Chence Shi},
  \bibinfo{person}{Zhiping Xiao}, \bibinfo{person}{Zhijian Duan},
  \bibinfo{person}{Yewen Xu}, \bibinfo{person}{Ming Zhang}, {and}
  \bibinfo{person}{Jian Tang}.} \bibinfo{year}{2019}\natexlab{}.
\newblock \showarticletitle{Autoint: Automatic feature interaction learning via
  self-attentive neural networks}. In \bibinfo{booktitle}{\emph{CIKM}}.
\newblock


\bibitem[\protect\citeauthoryear{Zha, Lai, Huang, Cao, Reddy, Vargas, Nguyen,
  Wei, Guo, and Hu}{Zha et~al\mbox{.}}{2021a}]%
        {zha2021rlcard}
\bibfield{author}{\bibinfo{person}{Daochen Zha}, \bibinfo{person}{Kwei-Herng
  Lai}, \bibinfo{person}{Songyi Huang}, \bibinfo{person}{Yuanpu Cao},
  \bibinfo{person}{Keerthana Reddy}, \bibinfo{person}{Juan Vargas},
  \bibinfo{person}{Alex Nguyen}, \bibinfo{person}{Ruzhe Wei},
  \bibinfo{person}{Junyu Guo}, {and} \bibinfo{person}{Xia Hu}.}
  \bibinfo{year}{2021}\natexlab{a}.
\newblock \showarticletitle{RLCard: a platform for reinforcement learning in
  card games}. In \bibinfo{booktitle}{\emph{IJCAI}}.
\newblock


\bibitem[\protect\citeauthoryear{Zha, Lai, Wan, and Hu}{Zha
  et~al\mbox{.}}{2020}]%
        {zha2020meta}
\bibfield{author}{\bibinfo{person}{Daochen Zha}, \bibinfo{person}{Kwei-Herng
  Lai}, \bibinfo{person}{Mingyang Wan}, {and} \bibinfo{person}{Xia Hu}.}
  \bibinfo{year}{2020}\natexlab{}.
\newblock \showarticletitle{Meta-AAD: Active anomaly detection with deep
  reinforcement learning}. In \bibinfo{booktitle}{\emph{ICDM}}.
\newblock


\bibitem[\protect\citeauthoryear{Zha, Lai, Zhou, and Hu}{Zha
  et~al\mbox{.}}{2019}]%
        {zha2019experience}
\bibfield{author}{\bibinfo{person}{Daochen Zha}, \bibinfo{person}{Kwei-Herng
  Lai}, \bibinfo{person}{Kaixiong Zhou}, {and} \bibinfo{person}{Xia Hu}.}
  \bibinfo{year}{2019}\natexlab{}.
\newblock \showarticletitle{Experience Replay Optimization}. In
  \bibinfo{booktitle}{\emph{IJCAI}}.
\newblock


\bibitem[\protect\citeauthoryear{Zha, Ma, Yuan, Hu, and Liu}{Zha
  et~al\mbox{.}}{2021b}]%
        {zha2020rank}
\bibfield{author}{\bibinfo{person}{Daochen Zha}, \bibinfo{person}{Wenye Ma},
  \bibinfo{person}{Lei Yuan}, \bibinfo{person}{Xia Hu}, {and}
  \bibinfo{person}{Ji Liu}.} \bibinfo{year}{2021}\natexlab{b}.
\newblock \showarticletitle{Rank the Episodes: A Simple Approach for
  Exploration in Procedurally-Generated Environments}. In
  \bibinfo{booktitle}{\emph{ICLR}}.
\newblock


\bibitem[\protect\citeauthoryear{Zha, Pervaiz~Bhat, Chen, Wang, Ding, Jain,
  Qazim~Bhat, Lai, Chen, et~al\mbox{.}}{Zha et~al\mbox{.}}{2022}]%
        {zha2021autovideo}
\bibfield{author}{\bibinfo{person}{Daochen Zha}, \bibinfo{person}{Zaid
  Pervaiz~Bhat}, \bibinfo{person}{Yi-Wei Chen}, \bibinfo{person}{Yicheng Wang},
  \bibinfo{person}{Sirui Ding}, \bibinfo{person}{Anmoll~Kumar Jain},
  \bibinfo{person}{Mohammad Qazim~Bhat}, \bibinfo{person}{Kwei-Herng Lai},
  \bibinfo{person}{Jiaben Chen}, {et~al\mbox{.}}}
  \bibinfo{year}{2022}\natexlab{}.
\newblock \showarticletitle{AutoVideo: An Automated Video Action Recognition
  System}. In \bibinfo{booktitle}{\emph{IJCAI}}.
\newblock


\bibitem[\protect\citeauthoryear{Zha, Xie, Ma, Zhang, Lian, Hu, and Liu}{Zha
  et~al\mbox{.}}{2021c}]%
        {zha2021douzero}
\bibfield{author}{\bibinfo{person}{Daochen Zha}, \bibinfo{person}{Jingru Xie},
  \bibinfo{person}{Wenye Ma}, \bibinfo{person}{Sheng Zhang},
  \bibinfo{person}{Xiangru Lian}, \bibinfo{person}{Xia Hu}, {and}
  \bibinfo{person}{Ji Liu}.} \bibinfo{year}{2021}\natexlab{c}.
\newblock \showarticletitle{DouZero: Mastering DouDizhu with Self-Play Deep
  Reinforcement Learning}. In \bibinfo{booktitle}{\emph{ICML}}.
\newblock


\bibitem[\protect\citeauthoryear{Zhang, Liu, Xie, Ktena, Tejani, Gupta, Myana,
  Dilipkumar, Paul, Ihara, et~al\mbox{.}}{Zhang et~al\mbox{.}}{2020}]%
        {zhang2020model}
\bibfield{author}{\bibinfo{person}{Caojin Zhang}, \bibinfo{person}{Yicun Liu},
  \bibinfo{person}{Yuanpu Xie}, \bibinfo{person}{Sofia~Ira Ktena},
  \bibinfo{person}{Alykhan Tejani}, \bibinfo{person}{Akshay Gupta},
  \bibinfo{person}{Pranay~Kumar Myana}, \bibinfo{person}{Deepak Dilipkumar},
  \bibinfo{person}{Suvadip Paul}, \bibinfo{person}{Ikuhiro Ihara},
  {et~al\mbox{.}}} \bibinfo{year}{2020}\natexlab{}.
\newblock \showarticletitle{Model size reduction using frequency based double
  hashing for recommender systems}. In \bibinfo{booktitle}{\emph{RecSys}}.
\newblock


\bibitem[\protect\citeauthoryear{Zhang, Yao, Sun, and Tay}{Zhang
  et~al\mbox{.}}{2019}]%
        {zhang2019deep}
\bibfield{author}{\bibinfo{person}{Shuai Zhang}, \bibinfo{person}{Lina Yao},
  \bibinfo{person}{Aixin Sun}, {and} \bibinfo{person}{Yi Tay}.}
  \bibinfo{year}{2019}\natexlab{}.
\newblock \showarticletitle{Deep learning based recommender system: A survey
  and new perspectives}.
\newblock \bibinfo{journal}{\emph{ACM Computing Surveys (CSUR)}}
  \bibinfo{volume}{52}, \bibinfo{number}{1} (\bibinfo{year}{2019}),
  \bibinfo{pages}{1--38}.
\newblock


\bibitem[\protect\citeauthoryear{Zhao, Xie, Jia, Qian, Ding, Sun, and Li}{Zhao
  et~al\mbox{.}}{2020b}]%
        {zhao2020distributed}
\bibfield{author}{\bibinfo{person}{Weijie Zhao}, \bibinfo{person}{Deping Xie},
  \bibinfo{person}{Ronglai Jia}, \bibinfo{person}{Yulei Qian},
  \bibinfo{person}{Ruiquan Ding}, \bibinfo{person}{Mingming Sun}, {and}
  \bibinfo{person}{Ping Li}.} \bibinfo{year}{2020}\natexlab{b}.
\newblock \showarticletitle{Distributed Hierarchical GPU Parameter Server for
  Massive Scale Deep Learning Ads Systems}. In
  \bibinfo{booktitle}{\emph{MLSys}}.
\newblock


\bibitem[\protect\citeauthoryear{Zhao, Wang, Chen, Zheng, Liu, and Tang}{Zhao
  et~al\mbox{.}}{2020a}]%
        {zhao2020autoemb}
\bibfield{author}{\bibinfo{person}{Xiangyu Zhao}, \bibinfo{person}{Chong Wang},
  \bibinfo{person}{Ming Chen}, \bibinfo{person}{Xudong Zheng},
  \bibinfo{person}{Xiaobing Liu}, {and} \bibinfo{person}{Jiliang Tang}.}
  \bibinfo{year}{2020}\natexlab{a}.
\newblock \showarticletitle{Autoemb: Automated embedding dimensionality search
  in streaming recommendations}. In \bibinfo{booktitle}{\emph{SIGIR}}.
\newblock


\bibitem[\protect\citeauthoryear{Zhou, Mou, Fan, Pi, Bian, Zhou, Zhu, and
  Gai}{Zhou et~al\mbox{.}}{2019}]%
        {zhou2019deep}
\bibfield{author}{\bibinfo{person}{Guorui Zhou}, \bibinfo{person}{Na Mou},
  \bibinfo{person}{Ying Fan}, \bibinfo{person}{Qi Pi}, \bibinfo{person}{Weijie
  Bian}, \bibinfo{person}{Chang Zhou}, \bibinfo{person}{Xiaoqiang Zhu}, {and}
  \bibinfo{person}{Kun Gai}.} \bibinfo{year}{2019}\natexlab{}.
\newblock \showarticletitle{Deep interest evolution network for click-through
  rate prediction}. In \bibinfo{booktitle}{\emph{AAAI}}.
\newblock


\bibitem[\protect\citeauthoryear{Zoph and Le}{Zoph and Le}{2017}]%
        {zoph2016neural}
\bibfield{author}{\bibinfo{person}{Barret Zoph} {and} \bibinfo{person}{Quoc~V
  Le}.} \bibinfo{year}{2017}\natexlab{}.
\newblock \showarticletitle{Neural architecture search with reinforcement
  learning}. In \bibinfo{booktitle}{\emph{ICLR}}.
\newblock


\end{thebibliography}

\newpage
\appendix

\begin{figure*}[]
  \centering
  \begin{subfigure}[b]{0.25\textwidth}
    \centering
    \includegraphics[width=0.9\textwidth]{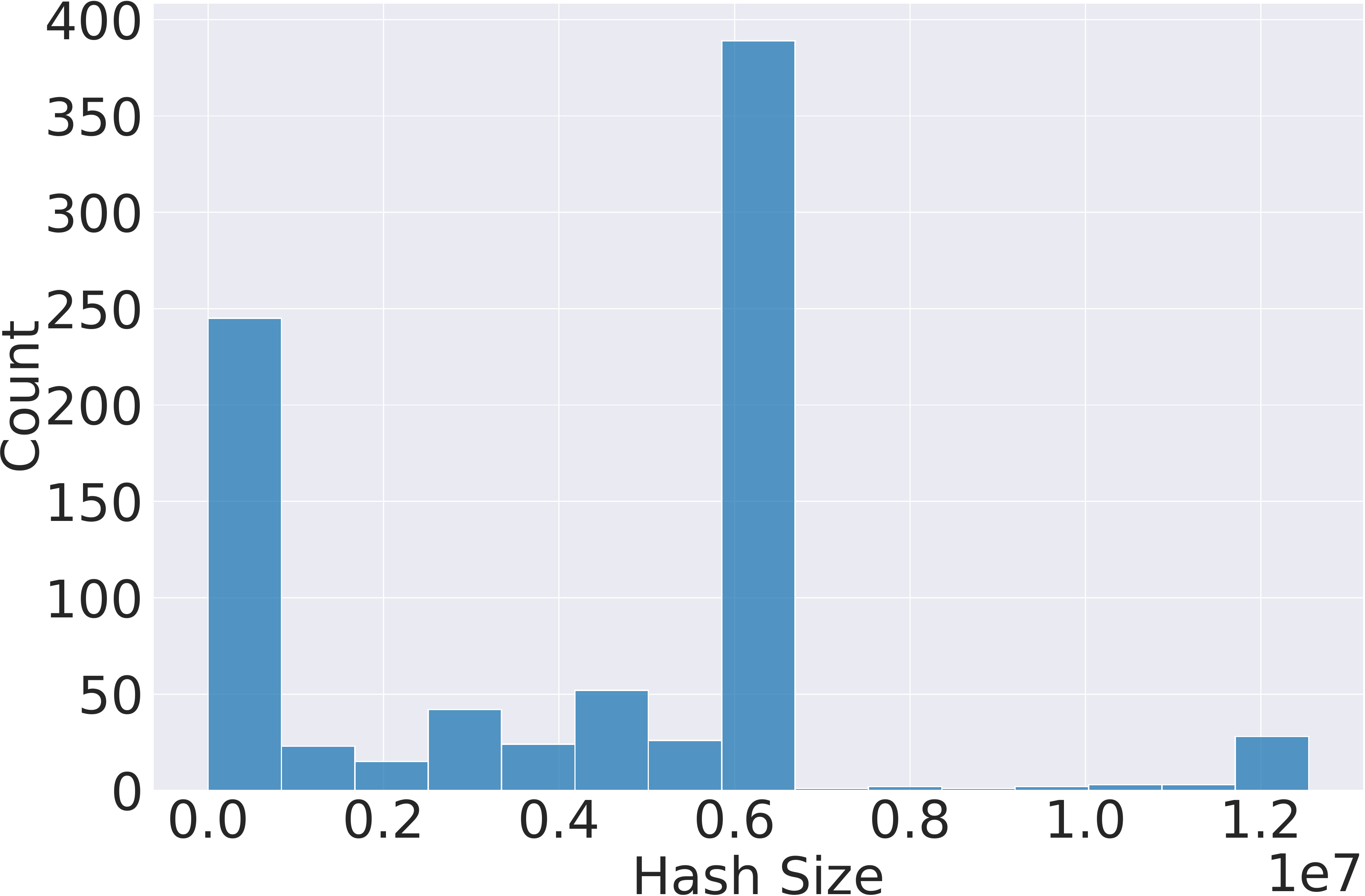}
    \vspace{-7pt}
    \caption{Hash size distribution}
  \end{subfigure}%
  \begin{subfigure}[b]{0.25\textwidth}
    \centering
    \includegraphics[width=0.9\textwidth]{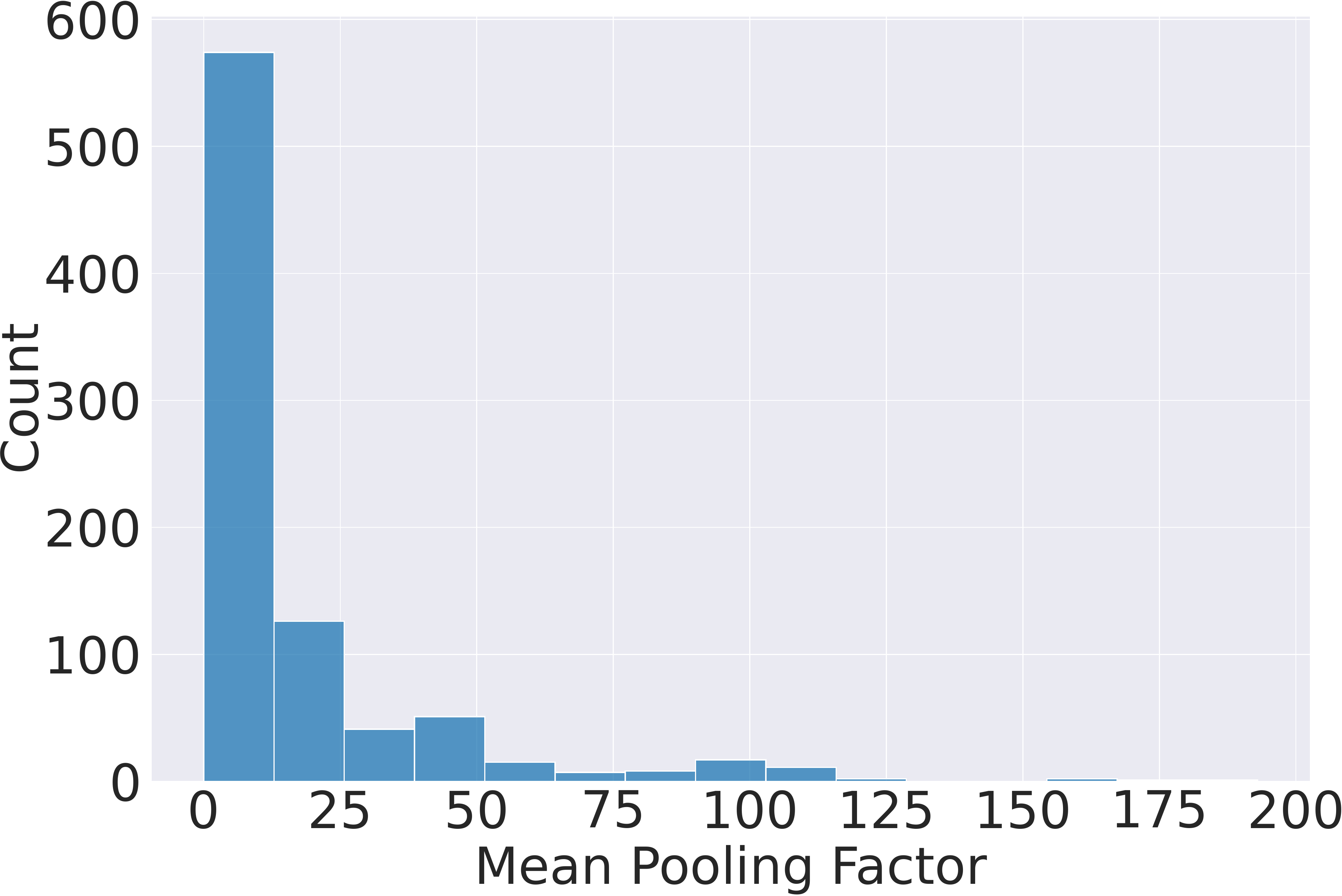}
    \vspace{-7pt}
    \caption{Pooling factor distribution}
  \end{subfigure}%
  \begin{subfigure}[b]{0.25\textwidth}
    \centering
    \includegraphics[width=0.9\textwidth]{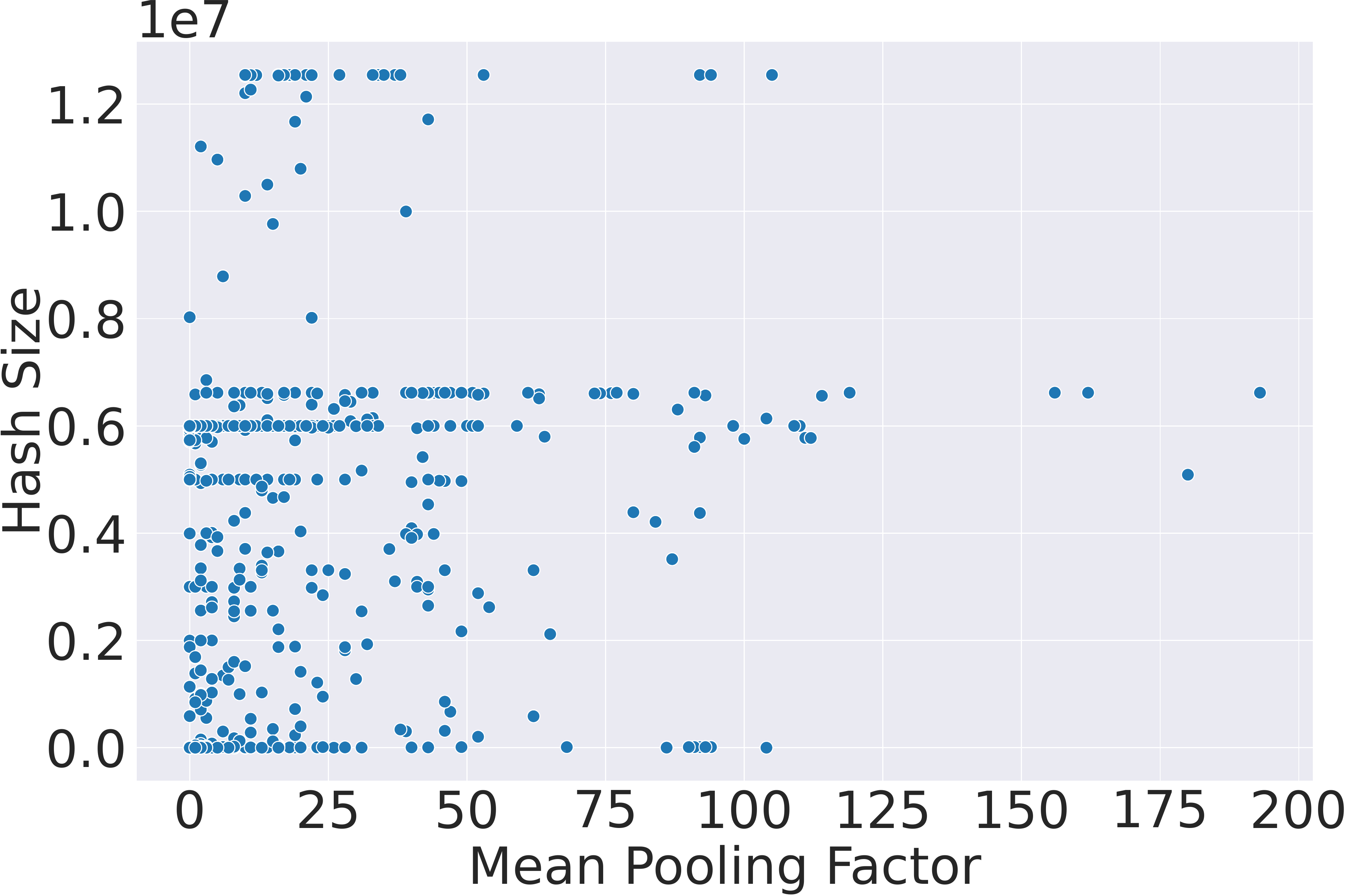}
    \vspace{-7pt}
    \caption{Hash size vs. pooling factor}
  \end{subfigure}%
  \begin{subfigure}[b]{0.25\textwidth}
    \centering
    \includegraphics[width=0.9\textwidth]{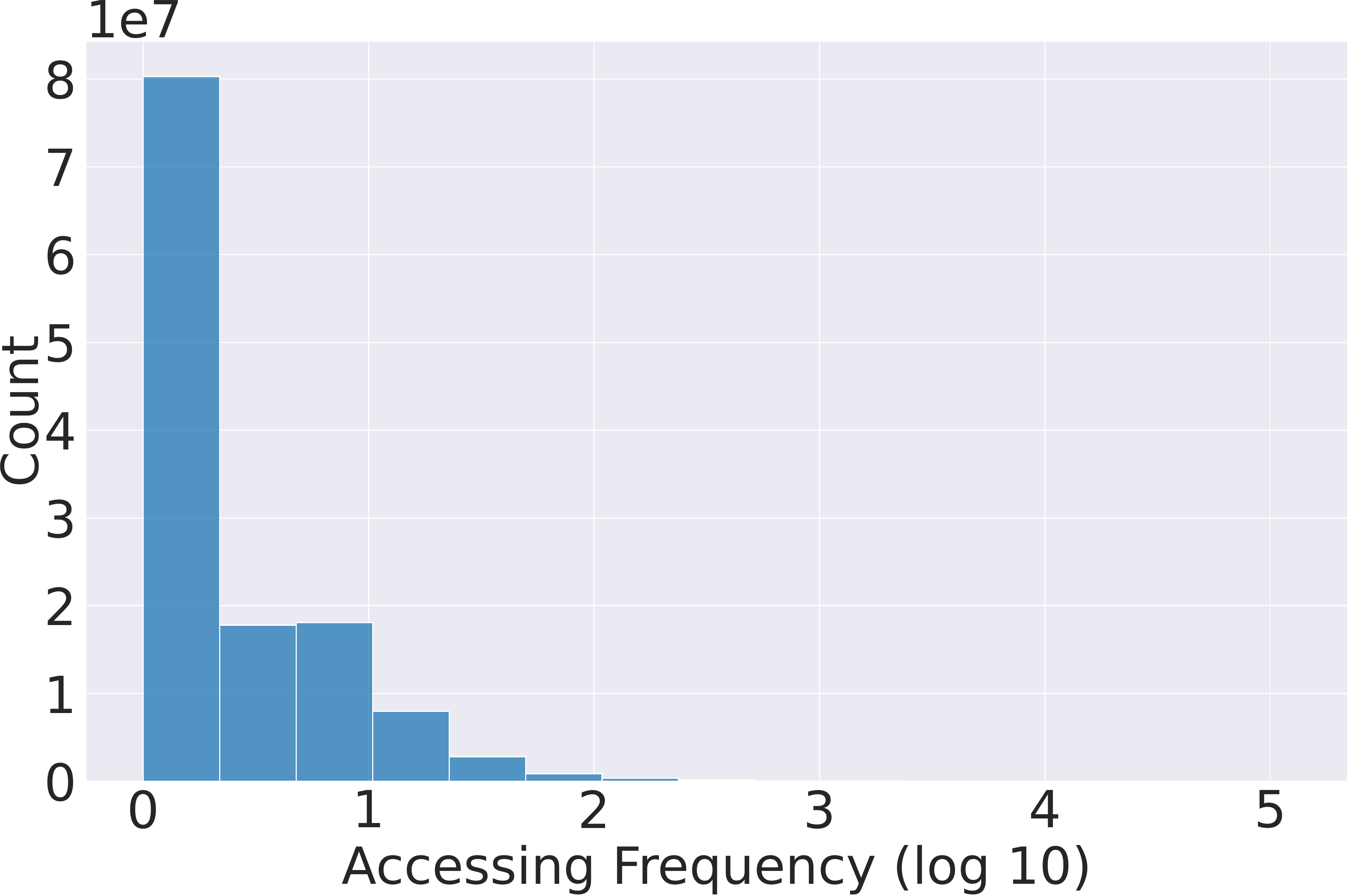}
    \vspace{-7pt}
    \caption{Indices frequency distribution}
  \end{subfigure}%
  \vspace{-8pt}
  \caption{Data distributions of MetaSyn.}
  \vspace{-8pt}
  \label{fig:datadistribtuion}
\end{figure*}

\section{Dataset Details}
\label{sec:appendix1}
We will not provide the details of MetaProd for data privacy and only discuss MetaSyn, which is open-sourced\footnote{\url{https://github.com/facebookresearch/dlrm_datasets}} and shares similar patterns to Meta production recommendation workloads.

\subsection{Data Format}
MetaSyn consists of three PyTorch tensors saved in a single file, including an indices tensor, an offsets tensor, and a length tensor. We denote them as \texttt{indices}, \texttt{offsets}, and \texttt{lengths}, respectively. \texttt{indices} is 1-dimensional, where each element is an integer index. The indices are ordered by \texttt{(table\_id, batch\_offset)}; that is, if we scan the tensor from the left to the right, we will first get a batch of indices for the first table, and then obtain a batch of indices for the second table, etc. \texttt{offsets} is also 1-dimensional and specifies the starting position and the ending position for one lookup. It is also ordered by \texttt{(table\_id, batch\_offset)}. For example, suppose the batch size is 65,536. Then \texttt{start = offsets[65536]} and \texttt{end = offsets[65537]} will specify the starting and ending positions of the first indices lookup in the second table. The indices tensor between the starting and ending positions, i.e., \texttt{indices[start:end]} correspond to the first instance in the batch and the second table. \texttt{lengths} is 2-dimensional and is of the shape \texttt{[num\_tables, batch\_size]}, where each element is the pooling factor of the corresponding indices lookup. \texttt{lengths} is provided for correctness validation since it can be inferred from the other two tensors. \texttt{indices} and \texttt{offsets} share the same data format with the batched embedding bag operator from FBGEMM and can be directly fed into it.

\subsection{Distributions}
We visualize the distribution of hash sizes, the distribution of mean pooling factors, the relation between the hash sizes and pooling factors, and the distribution of indices accessing frequency of MetaSyn in Figure~\ref{fig:datadistribtuion}. We make the following observations: 1) the hash sizes for most tables are on the scale of millions, while some can reach tens of millions. 2) the pooling factor generally follows a power-law distribution, where the majority is less than 50, while some can be as large as 200. 3) there is no clear relationship between the hash size and pooling factor; 4) most of the indices are accessed less than ten times, while some of them can reach $10^5$. The highly diverse table characteristics and indices patterns will easily result in imbalances if not carefully partitioning the tables. Specifically, the costs of the tables will also be diverse, i.e., some tables will have extremely high costs while some others could have very low costs. As a result, if we do not shard the tables carefully, some tables with high costs can be easily put into the same shard, resulting in a very high cost for the shard.

\subsection{Data Processing}

Recall that the 856 tables in MetaSyn will serve as the table pool in our experiments. For the ease of use, we separate \texttt{indices} and \texttt{offsets} into a list of \texttt{indices} and a list of \texttt{offsets}, respectively, where each element corresponds to one table. The offsets for each table will be re-indexed starting from 0. In training, we merge indices and the offsets of the selected tables to construct the input of the operator. For the embedding tables, we randomly choose the table dimension from $\{16,32\}$ since the MetaSyn does not provide this information. We purposely make the dimensions small to facilitate reproducibility on GPUs with small memory. Note that, for MetaProd, the table dimension is specified by the production model and is larger than MetaSyn. For the parameters of embedding tables, we randomly initialize them with fp16 precision.

\section{Model Details}
\label{sec:appendix2}

\begin{algorithm}[t]
\caption{\texttt{benchmark\_op}}
\label{alg:2}
\setlength{\intextsep}{0pt} 
\begin{algorithmic}[1]
\STATE \textbf{Input:} A PyTorch operator, the arguments of the operator
\STATE Feed random tensors to GPU to clear the cache
\STATE Initialize the start and end CUDA events for time measuring
\STATE Run the forward and backward passes of the operator
\STATE Collect latency from the CUDA events

\STATE Return the latency
\end{algorithmic}
\end{algorithm}

\begin{algorithm}[t]
\caption{Micro-Benchmark}
\label{alg:3}
\setlength{\intextsep}{0pt} 
\begin{algorithmic}[1]
\STATE \textbf{Input:} A PyTorch operator, the arguments of the operator, the number of warmup iterations $W$, the number of measuring iterations $B$, the number of removed highest/lowest costs $R$

\FOR{iteration = 1, 2, ... W}
    \STATE Call \texttt{benchmark\_op} with the operator and the arguments
\ENDFOR

\STATE Initialize a Python list \texttt{costs} to store the costs
\FOR{iteration = 1, 2, ... B}
    \STATE Call \texttt{benchmark\_op} and append the result to \texttt{costs}
\ENDFOR
\STATE Sort the \texttt{costs} and remove the $R$ highest/lowest costs.
\STATE Return the mean of the \texttt{costs}
\end{algorithmic}
\end{algorithm}

\subsection{Details of Micro-Benchmark}
\label{sec:appendix21}
Precisely measuring the kernel time of the embedding operator requires non-trivial engineering efforts because of the warmup overhead of the GPU, caching mechanism, and variance. Specifically, naively running an operator multiple times and using the mean latency as the cost will not result in an accurate estimation. First, the warmup stage of the CUDA devices will cause significant overhead, leading to a higher estimation of the cost. Second, the L1 and L2 caches will cache the tensors in the previous iterations and result in a lower estimation of the cost. Third, some certain anomalous runs could have very high or low latency due to variance. However, the mean latency will be sensitive to anomalies. Fortunately, we found that some engineering tricks can well tackle the above problems to enable a precise measurement of the latency. The overall procedure is summarized in Algorithm~\ref{alg:3} with an inner function in Algorithm~\ref{alg:2}. Before actually measuring the time, we first run several warmup iterations to warm up the GPU (line 2-3 of Algorithm~\ref{alg:3}). For each run, we first clear the cache to remove the impact of caching (line 2 of Algorithm~\ref{alg:2}). To tackle the anomalies, we remove the highest and the lowest costs and return the mean value of the remaining costs. We empirically set $W=5$, $B=10$, and $R=2$. We have performed sanity checks and concluded that the time collected from the micro-benchmark is consistent with the kernel time obtained by profiling. In essence, the micro-benchmark is general and can be applied to other operators as well. Thus, we have deployed it in the production environment to support general micro-benchmarking. It is also open-sourced under PARAM Benchmarks\footnote{\url{https://github.com/facebookresearch/param}}.

\begin{table}[]
\centering
\caption{Mean absolute error (MAE) and mean squared error (MSE) of the cost model on 100 randomly sampled 10 tables following the unseen tables setting in Section~\ref{sec:45}, where the training tables are the first half of the pool, and the testing tables are the second half. The prediction error of the multi-table cost is around 1.3 milliseconds on the training tables and 3 milliseconds on the unseen tables.}
\vspace{-8pt}
\label{tbl:costmodel}
\footnotesize
\begin{tabular}{c|c||c|c}
 \toprule
 \textbf{MAE (training)} & \textbf{MSE (training)} & \textbf{MAE (testing)} & \textbf{MSE (testing)} \\
 \midrule
 1.321 & 3.408 & 3.061 & 10.835\\

 \bottomrule
\end{tabular}
\vspace{-10pt}
\end{table}

\subsection{Details of Features}
\label{sec:appendix22}
The features are important for learning. Recall that we have used the following features in the cost model and the RL policy: table dimension, hash size, pooling factor, table size, indices distributions, and step-aware feature. We elaborate on these features below.
\begin{itemize}
    \item \textbf{Table dimension:} it is the dimension of each embedding vector in the table. This feature is normalized to have a mean of 0 and a standard deviation of 1.
    \item \textbf{Hash size:} it is the number of rows in a table, which is determined by the feature cardinality. This feature is also normalized.
    \item \textbf{Pooling Factor:} it is obtained by dividing the total number indices with the batch size. This feature is also normalized.
    \item \textbf{Table size:} it is obtained by calculating the parameter size in GBs. We do not normalize this feature.
    \item \textbf{Indices distributions:} Since the majority of the indices are visited very few times, we divide frequencies into several bins, where the size of each bin grows exponentially. Specifically, we use the following 17 bins: $(0, 1]$, $(1, 2]$, $(2, 4]$, $(4, 8]$, $(8, 16]$, $(16, 32]$, $(32, 64]$, $(64, 128]$, $(128, 256]$, $(256, 512]$, $(512,$ $1024]$, $(1024, 2048]$, $(2048, 4096]$, $(4096, 8192]$, $(8192, 16384]$, $(16384, 32768]$, and $(32768, \infty)$. We count the number of times each index appears in a batch of indices and put the count into the corresponding bin. Then we calculate the ratio for each bin, which results in 17 features. For example, the first feature is the ratio of the indices for the bin $(0, 1]$. The fourth feature is the ratio of the indices for the bin $(4, 8]$.
    \item \textbf{Step-aware feature:} it is defined as the the ratio of the tables that have already been assigned.
    
\end{itemize}

\section{Implantation Details}
\label{sec:appendix3}

\subsection{Neural Architecture}
\textbf{Architecture of the cost model.} We use a two-layer MLP to obtain the representation for every single table. The input dimension is 21, which is the number of table features. The hidden dimension is 128. The output size is 32; that is, the size of a single table representation is 32. For multi-table cases, we sum the table representations to obtain multi-table representation, whose size is also 32. Finally, we use another two-layer MLP to produce the multi-table cost, where the hidden dimension is 64, and the output size is 1.

\textbf{Architecture of the policy-value network.} We first use one-layer MLP to map the state-aware feature to a 32-dimensional representation. Then we concatenate the single table 32-dimensional representation (shared with the cost model) of the upcoming table to obtain a 64-dimensional state representation. The state representation is processed by a two-layer LSTM with a hidden size 64. For actions, we similar obtain a multi-table representation for each shard. We use a one-layer MLP to map the shard cost to a 32-dimensional cost representation. The action representation is obtained by concatenating the multi-table representation and the cost representation, and is 64-dimensional. For the policy head, we concatenate the state representation, the action representation, and their dot-product and use a four-layer MLP with size 128-128-64--1 to generate the action confidence. Then SoftMax is applied to obtain the probabilities. For the value head, we use a two-layer MLP with a size 64-1 to produce the state value.

\subsection{Hyperparameters Configuration}
\label{sec:appendix31}
We summarize all the hyperparameters of AutoShard below.
\begin{itemize}
    \item \textbf{Cost model training}: batch size $B_2=512$,  the number of update iteration $I=20$, the buffer size is 5000.
    \item \textbf{RL training}: batch size $B_1=8$, number of data collection steps $T=100$, number of learning threads is 1, entropy weight is 0.001, baseline weight is 0.5, discounting factor is 1.0, gradient is clipped with threshold 40, and the total number of training steps is 100,000.
    \item \textbf{Optimizer}: We use Adam optimizer with a learning rate of 0.001. The other hyperparameters are kept as default.
\end{itemize}

\subsection{Hardware}
\label{sec:appendix32}
For MetaSyn, we conduct all the experiments on a server with 48 Intel(R) Xeon(R) Silver 4116 CPU @ 2.10GHz processors, 188 GB memory, and four NVIDIA GeForce RTX 2080 Ti GPUs. For MetaProd, the server has a similar hardware environment but with NVIDIA V100 GPUs.

\end{document}